\documentclass[10pt,twocolumn,letterpaper]{article}

\usepackage[pagenumbers]{iccv} 

\usepackage{amsmath,amsfonts,bm}

\def\eqref#1{equation~\ref{#1}}

\def\1{\bm{1}}

\DeclareMathAlphabet{\mathsfit}{\encodingdefault}{\sfdefault}{m}{sl}
\SetMathAlphabet{\mathsfit}{bold}{\encodingdefault}{\sfdefault}{bx}{n}

\definecolor{iccvblue}{rgb}{0.21,0.49,0.74}
\usepackage[pagebackref,breaklinks,colorlinks,allcolors=iccvblue]{hyperref}

\usepackage{url}            
\usepackage{booktabs}       
\usepackage{amsfonts}       
\usepackage{nicefrac}       
\usepackage{microtype}      

\usepackage{multirow}
\usepackage{amsmath}

\usepackage{graphicx}
\usepackage{svg}
\usepackage{scalefnt}
\usepackage{chngcntr}
\usepackage{amssymb}
\usepackage{pifont}
\usepackage{titletoc} 
\usepackage{wrapfig}
\usepackage{caption}
\usepackage[table]{xcolor}
\usepackage[table,xcdraw]{xcolor}
\usepackage{amssymb}
\usepackage{marvosym}  
\def\paperID{13400} 
\def\confName{ICCV}
\def\confYear{2025}

\title{UIPro: Unleashing Superior Interaction Capability For GUI Agents}

\author{
  Hongxin Li$^{1,2,3,7}$\quad 
  Jingran Su$^{5}$ \quad
  Jingfan Chen$^{5}$ \quad 
  Zheng Ju$^{1,2,3}$ \quad
  Yuntao Chen$^{4}$\textsuperscript{\Letter} \quad 
  Qing Li$^{5}$\\
  Zhaoxiang Zhang$^{1,2,3,6}$\textsuperscript{\Letter} \quad
  \\ 
$^1$University of Chinese Academy of Sciences (UCAS) \\
$^2$New Laboratory of Pattern Recognition (NLPR), CASIA \\
$^3$State Key Laboratory of Multimodal Artificial Intelligence Systems (MAIS), CASIA \\
$^4$Hong Kong Institute of Science \& Innovation, CASIA \\
$^5$PolyU \quad
$^6$Shanghai Artificial Intelligence Laboratory \quad
$^7$StepFun \\
  \small{Code: \url{https://github.com/ZJULiHongxin/UIPro}}
}

\newcommand{\xmark}{\ding{55}}
\newcommand{\cross}{\textcolor{red}{\xmark}}
\colorlet{Mycolor1}{green!100}
\newcommand{\cmark}{\textcolor{Mycolor1}{\ding{51}}}

\newcommand{\methodname}{UIPro}

\begin{document}
\maketitle

\newcommand{\myfootnote}[1]{%
  \begingroup
  \renewcommand\thefootnote{}\footnotetext{#1}%
  \endgroup
}

\myfootnote{\textsuperscript{\Letter} Equally advising corresponding authors. E-mails: zhaoxiang.zhang@ia.ac.cn, chenyuntao08@gmail.com.}

\begin{abstract}
Building autonomous agents that perceive and operate graphical user interfaces (GUIs) like humans has long been a vision in the field of artificial intelligence. Central to these agents is the capability for GUI interaction, which involves GUI understanding and planning capabilities. Existing methods have tried developing GUI agents based on the multi-modal comprehension ability of vision-language models (VLMs). However, the limited scenario, insufficient size, and heterogeneous action spaces hinder the progress of building generalist GUI agents. To resolve these issues, this paper proposes \textbf{UIPro}, a novel generalist GUI agent trained with extensive multi-platform and multi-task GUI interaction data, coupled with a unified action space. We first curate a comprehensive dataset encompassing 20.6 million GUI understanding tasks to pre-train UIPro, granting it a strong GUI grounding capability, which is key to downstream GUI agent tasks. Subsequently, we establish a unified action space to harmonize heterogeneous GUI agent task datasets and produce a merged dataset to foster the action prediction ability of UIPro via continued fine-tuning. Experimental results demonstrate UIPro's superior performance across multiple GUI task benchmarks on various platforms, highlighting the effectiveness of our approach.
\end{abstract}    
\section{Introduction}

\begin{figure}[t]
    \centering
    \includegraphics[width=0.92\linewidth]{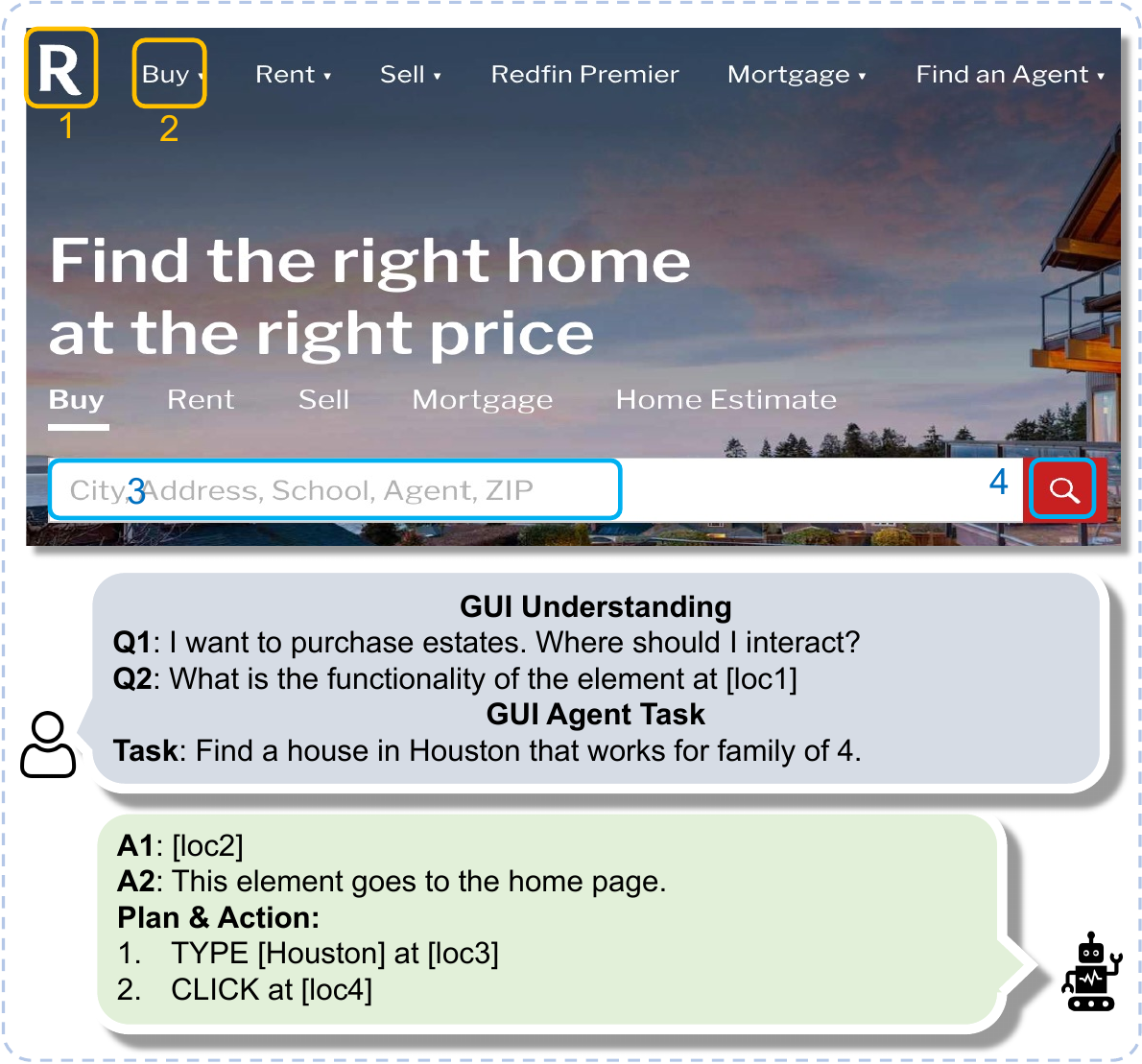}
    \caption{Examples of GUI interaction by a GUI agent.}
    \label{fig: agent demo}
\end{figure}

The concept of autonomous GUI agents capable of clicking, typing, and scrolling on behalf of humans as personal assistants is an enticing prospect (Fig.~\ref{fig: agent demo}).
Imagine a GUI agent navigating the Internet to perform daily tasks such as using search engines and managing emails, as well as more complex activities like comparing prices across e-commerce platforms and collecting the latest news on stock markets.

To achieve proficient GUI interaction, an agent relies on two fundamental capabilities. Firstly, a foundational visual understanding of GUI elements is essential. This includes the ability to recognize and interpret various components like buttons, text fields, and images, which are integral to navigating and manipulating GUIs effectively. The second capability involves both planning and executing tasks in alignment with user goals, requiring the agent to efficiently translate plans into actionable steps.

Although existing methods~\cite{cheng2024seeclick,hong2023cogagent,you2024ferretui,GUICourse} have sought to enhance GUI interaction by leveraging the multi-modal comprehension abilities of vision-language models (VLMs), significant challenges persist: (a) \textbf{Data scale.} Effective development of generalist GUI agents demands large-scale data since the complete advantages of large-scale training typically do not manifest at smaller scales~\cite{wei2022emergent,pi0}. Unfortunately, current GUI interaction datasets~\cite{Li2020WidgetCG,Burns2022ADF,baechler2024screenai,liu2024visualwebbench,WebUI,deng2024mind2web} often lack sufficient size and scenario diversity, hindering the development of generalist GUI agents. (b) \textbf{Training recipe} is also critical, as the recent advancements in large language and vision models have largely depended on sophisticated strategies for data curation~\cite{rlhf,liu2024llavanext}. Existing GUI interaction datasets typically adopt various formats (e.g., heterogeneous action spaces), which complicates data curation and reduces the effectiveness of training generalist GUI agents.

In this paper, we tackle the aforementioned issues by developing \textbf{UIPro}, a generalist GUI agent trained with massive multi-platform multi-task GUI interaction data and unified action spaces. Concretely, to lay the foundation of GUI interaction, we first curate a large-scale GUI-understanding-oriented dataset by annotating and curating data from multiple GUI platforms, including web browsers, Android phones, and iPads of various types and resolutions. We also spot the astonishingly high noise level in the raw GUI sources and release a systematic denoising procedure to reduce the burden of data cleaning in future research. After cleaning, we curate \textbf{20.6M} GUI understanding tasks associated with \textbf{2.5M} unique screenshots, contributing the largest GUI understanding data collection to the community. This dataset is used to pre-train UIPro to develop a strong GUI understanding, especially element grounding capability, which is the key to downstream GUI agent tasks.

To cultivate UIPro's action prediction ability after pre-training, we propose to better combine multiple heterogeneous GUI agent task datasets with a unified action space. This unified space defines an action superset that accommodates different action definitions used in various data sources. Specifically, we resolve action name conflicts, utilize action arguments suitable for various downstream test environments, and finally format all action ground truths in a unified JSON format. As different platforms usually share similar design principles for human-computer interaction, the unified action space that abstracts platform distinctions helps to develop adaptable GUI agents.

After fine-tuning with unified agent task data, UIPro achieves superior performance on multiple GUI interaction benchmarks on different platforms. Ablation studies also justify that the 20.6M GUI understanding data and the proposed unified action space both notably benefit GUI interaction capabilities.

The main contributions of our work are three-fold:

1. We curate a massive and clean GUI understanding dataset with 20.6M task samples on multiple platforms to imbue strong GUI grounding capability into GUI agents.

2. We propose to unify the action space of heterogeneous GUI agent task datasets. This unified action space can better integrate multiple data sources to enhance the GUI interaction capability of GUI agents.

3. Undergoing training with the curated GUI understanding dataset and unified agent task dataset, we build UIPro, an advanced GUI agent that achieves superior agent task performance on multiple platforms.
\section{Related Works}

\subsection{Existing GUI Datasets}
Compared with natural image datasets~\citep{Russakovsky2014ImageNetLS,LAION5B}, GUI datasets have received less attention. Some works have developed datasets for mobile GUIs~\citep{Wang2021Screen2WordsAM,Li2020WidgetCG,Li2020MappingNL,Bai2021UIBertLG,Burns2022ADF,chen2020wireframe} centered on the RICO dataset~\citep{deka2017rico}, which contains 72K Android app screenshots. Notable examples include Widget Captioning~\citep{Li2020WidgetCG}, which examines the captions and linguistic features of UI elements, and RICOSCA~\citep{Li2020MappingNL}, which maps single-step instructions to corresponding UI elements. MobileViews~\cite{mobileviews} utilizes chip clusters to parallelize GUI data collection, providing a huge dataset covering 20k Apps. Some works focus solely on web scenarios: WebUI~\cite{WebUI} crawls 400k rendered webpages associated with automatically extracted metadata used for enhancing GUI visual understanding. GUICourse~\cite{GUICourse} renders URLs from the \textit{C4}~\cite{2020t5} and generates element-level understanding, GUI Q\&A, and action prediction tasks for building GUI agents. To expand diversity, ~\cite{baechler2024screenai} and \cite{hong2023cogagent} curate huge datasets at the scale of hundreds of millions, but these datasets have not been publicized. Despite the smaller scale, several works~\cite{cheng2024seeclick,multiUI,uground,osatlas} have collected multi-platform datasets, helping to push the boundary of GUI agents. Our work also contributes a new large-scale GUI understanding dataset by cleaning, annotating, and curating GUI data on multiple platforms.

\subsection{GUI Agents}

Existing works~\cite{nakano2021webgpt,gur2023real, you2024ferretui,  cheng2024seeclick, zhang2023appagent,aguvis,uitars} have explored GUI agents leveraging the emergent capabilities of LLMs and VLMs to tackle long-horizon tasks.
Some studies focus on text-based environments~\cite{yao2022webshop,nakano2021webgpt,deng2024mind2web,gur2023real,zheng2023synapse}, with minimal interaction in more complex, dynamic settings, such as visual GUI environments. Operating GUIs like humans presents significant challenges for autonomous agents due to the need for pixel-level control within a complex action space. To address these challenges, recent studies have emphasized visually grounded, pixel-level GUI agents~\cite{hong2023cogagent,bai2024digirl, you2024ferretui, shaw2023pixels, cheng2024seeclick, zhang2023appagent, uground, osatlas,aguvis,uitars}. For instance, Pixel2Act~\cite{shaw2023pixels} and WebGUM~\cite{furuta2023multimodal} concentrate on the web domain where they develop models that can predict actions on the HTML elements with visual information. Ferret-UI~\cite{you2024ferretui}, SeeClick~\cite{cheng2024seeclick}, CogAgent~\cite{hong2023cogagent}, and OS-ATLAS~\cite{osatlas} curate large-scale data to train open-source multimodal models as GUI agents. Additionally, some works~\cite{OmniParser, uground} also explore combining proprietary foundation models (e.g., GPT-4~\cite{openai2024gpt4} and PaLM2~\cite{anil2023palm2technicalreport}) as task decomposers while training VLMs as low-level action executors. However, the generalizability of these works is often insufficient trajectory data and inconsistent action spaces. To address this, we propose a unified action space to integrate multiple data sources, enhancing GUI interaction capabilities.

\section{UIPro: Advanced GUI Agent}

\begin{figure*}[t]
    \centering
    \includegraphics[width=0.93\linewidth]{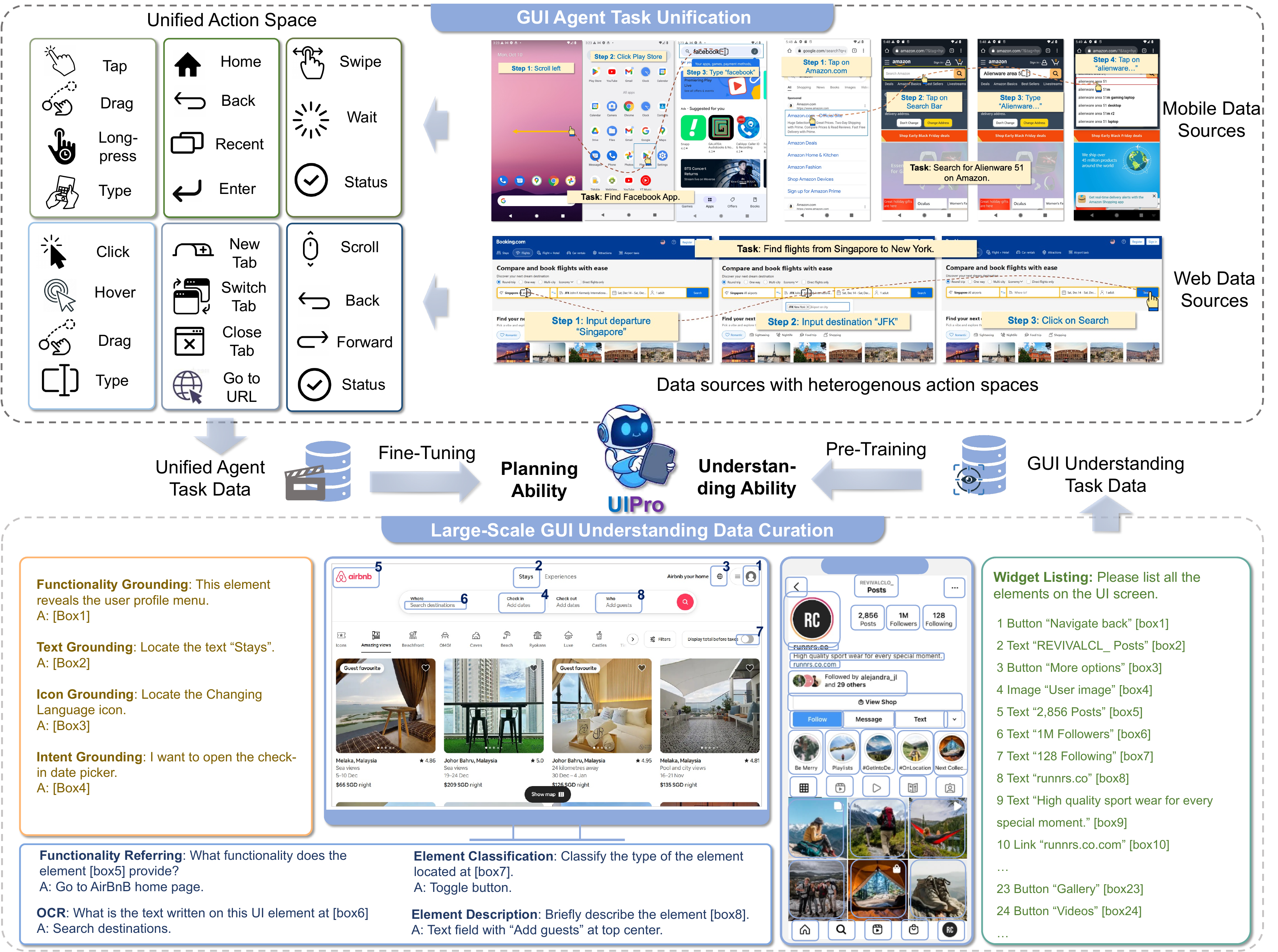}
    \caption{We develop \textbf{UIPro}, an advanced GUI agent capable of interacting with GUIs given user tasks. We first curate a large-scale GUI understanding dataset containing diverse tasks used to pre-train UIPro. Then, we merge heterogeneous GUI agent task data with unified action spaces to fine-tune UIPro, granting UIPro superior planning ability, }
    \label{fig: main figure}
\end{figure*}

This section introduces UIPro, an advanced GUI agent built upon massive GUI interaction data curated by us (Fig.~\ref{fig: main figure}).
\subsection{Model Architecture}
UIPro is designed to perform GUI interaction by receiving user tasks as input and then locating target elements (GUI element grounding) and predicting actions advancing towards task completion (GUI agent task). UIPro adopts the popular architecture used by recent VLMs, such as LLaVA~\citep{liu2023llava}. This architecture combines a pre-trained visual backbone $f_{\phi}$  (e.g., ViT~\citep{dosovitskiy2021vit}) and a large language model (e.g., Llama~\cite{touvron2023llama}) to build a model capable of processing both textual and visual inputs.
We build UIPro on two base models capable of processing high-resolution GUI screenshot images: \textbf{UIPro-SLiME} trained from a tabula rasa SLiME-Gemma-2B~\cite{slime}, and \textbf{UIPro-Qwen2VL} fine-tuned from Qwen2-VL-7B-Instruct~\cite{qwen2vl}.

\subsection{GUI Interaction Data Construction}

We curate massive GUI understanding and agent task data to grant UIPro strong GUI interaction performance.
\subsubsection{Curating GUI Understanding Data}
\label{sec:gnd data construc}
\begin{table}[t]
\scriptsize
\centering
\resizebox{\linewidth}{!}{%

\begin{tabular}{@{}ccccccccc@{}}
\toprule
Dataset & UI Type & \#Samples & \#Task Types  & Open-source\\ \midrule
Wid. Cap.~\citep{Li2020WidgetCG} & Mobile &  163k & 1 & \cmark \\
RICOSCA~\citep{Li2020MappingNL} & Mobile &  295k & 1 & \cmark \\
RefExp~\citep{Bai2021UIBertLG} & Mobile &  390k & 1 &  \cmark \\
SeeClick~\citep{cheng2024seeclick} & Web, Mobile &  5.3M & 6 &  \cmark \\
CogAgent~\citep{hong2023cogagent} & Web &   $>$247M & 5  & \cross \\
Ferret-UI~\citep{you2024ferretui} & Mobile &   250k & 11 & \cross \\
ScreenAI~\citep{baechler2024screenai} & Mobile &   421M & 4 &  \cross \\
UGround~\citep{uground} & Web, Mobile &  10M & 5 & \cmark \\
OS-ATLAS~\citep{osatlas} & Web, Mobile, Desktop &  13.6M & 4 & \cmark \\

GUICourse~\citep{GUICourse} & Web, Mobile &   10.8M & 3 & \cmark \\
\methodname{} (ours) & Web, Mobile &  20.6M & \textbf{13} & \cmark \\ \bottomrule
\end{tabular}
}
\vspace{-1mm}

\caption{Comparing large-scale datasets tailored to fostering GUI understanding ability.}

\label{tab:data comparison}
\end{table}

To imbue GUI understanding, especially \textbf{\textit{grounding}} capability, into UIPro, we first curate a super-large dataset by annotating crawled GUI data and generating diverse GUI grounding tasks. Each grounding task is represented by a \textit{\textless screenshot, referring expression (RE), coordinates\textgreater} triplet. We generate various REs for GUI elements:
\textbf{(a) Element Description} describes visual appearances, element types, and positions of elements, aiming to establish a fundamental GUI understanding capability. For example, a house-shape element might be described as ``a navigating-home button at the top left''. We also extract the displayed texts for pure-text elements and icon classes (e.g., \textit{Home icon} and \textit{Instagram icon}) for iconic ones as their REs.
\textbf{(b) User Intent} describes how a user intends to interact with a specific element~\cite{Burns2022ADF}, such as ``focus on the Password textbox'' for a password input field. The aforementioned two types of REs are derived from the element properties in GUI source code, such as HTML or Android view hierarchies.
\textbf{(c) Contextual Functionality} describes the interactive affordance of GUI elements~\cite{li2025autoguiscalingguigrounding}. For example, an element shaped like an upward arrow might be annotated with ``This element enables users to share content with others''. This RE type helps models comprehend the functional semantics of elements, complementing the textual and appearance-level semantics provided by the former two types. We generate this RE type following~\cite{li2025autoguiscalingguigrounding}.

Using the triplets, we generate massive GUI grounding tasks, including \textit{funcgnd}, \textit{elemgnd}, \textit{textgnd}, \textit{icongnd}, and \textit{intentgnd}) based on the functionality annotations, element descriptions, displayed texts, icon classes, and user intents, respectively. Following~\cite{cheng2024seeclick} and ~\cite{hong2023cogagent}, we also produce dual referring tasks (e.g., \textit{funcref}, \textit{elemref}, \textit{OCR}, \textit{iconref}) that prompt the model to generate referring expressions for specific element locations.
To further enhance understanding capability, we follow Ferret-UI~\cite{you2024ferretui} to also include \textit{widget listing}, \textit{GUI captioning}, and \textit{GUI Q\&A} tasks.

To expand scenario diversity, the used GUI sources include multi-resolution webpages from Common Crawl, mobile UIs from various Android device types, and publicized raw GUI collections (details in the Appendix).

\subsubsection{Unifying GUI Agent Task Data}
\label{sec:unify agent data}

After pretraining UIPro with the GUI understanding data, we fine-tune it on downstream GUI agent tasks. Although various GUI trajectory datasets exist, each offers limited training trajectories and employs different action spaces with inconsistent implementations. For instance, AITW~\cite{rawles2023android} defines \textit{swipe} as \verb|DUAL_POINT(start, end)| while AndroidControl~\cite{androidcontrol} uses \verb|scroll(direction)|. Mind2Web~\cite{deng2024mind2web} requires a GUI agent to output the text box coordinates for a typing action, while GUIAct~\cite{GUICourse} and WebLINX~\cite{weblinx} do not support this coordinate argument for typing. Moreover, swipe and scroll are not used consistently across the existing datasets. These conflicting definitions complicate combining multiple sources for multi-task fine-tuning.

To handle this problem, we merge different sources of GUI agent tasks by proposing a unified action space, which employs general action definitions as a superset over heterogeneous ones. For example, we define \textit{swipe} as \verb|swipe(start, direction, distance)| to represent how users start the swipe at a specific point on the screen, move their finger in a certain direction, and cover a certain distance to complete the swipe gesture. This definition accommodates the swipe usages of multiple datasets, including AITW~\cite{rawles2023android} and AndroidControl~\cite{androidcontrol}. We design three unified action spaces for mobile devices, web browsers, and desktop environments, respectively, representing three different embodiments of UIPro. For instance, the mobile action space incorporates \verb|tap|, \verb|long_press|, \verb|drag|, \verb|input_text|, \verb|navigate_home/back/recent|, \verb|press_enter|, \verb|swipe|, \verb|wait|, \verb|status_complete/infeasible| (Full definitions in Sec.~\ref{sec:supp: act space def} of Appendix). This action unification enhances UIPro's interoperability and facilitates the integration of diverse data sources for more effective multi-task learning. As training data in mobile and web scenarios are significantly richer than those in desktop environments (e.g., Windows and MacOS), we mainly merge datasets in mobile and web scenarios to justify the efficacy of the proposed action space unification.

\subsubsection{Systematic Denoising Procedure}
\label{sec: denoise}
GUI data may contain noise due to GUI design defects\footnote{A recent survey (\url{https://webaim.org/projects/million/}) revealed that 95.9\% of home pages contained accessibility errors, averaging 56.8 errors per page.}. Training samples generated based on these noisy elements will likely compromise the performance of the trained model, as shown in our experiments. To tackle this issue, we inspect the used GUI data sources, summarize invalid element types, and contribute a denoising procedure useful for the community. The checking items of the procedure include detecting blank elements according to the standard deviation of region color, employing OCR tools\footnote{https://pypi.org/project/pytesseract/} to remove invisible elements, and removing elements with invalid bounding boxes (the full procedure is detailed in Sec.~\ref{sec:supp:denoise pipeline} of Appendix).
After checking multiple data sources, we surprisingly found that the noise ratio is unignorable, with one source~\cite{chen2020wireframe} reaching $29.0\%$, as shown in Tab.~\ref{tab:noise ratio}. We use this denoising procedure to ensure our dataset is generated from clean elements.
\begin{figure}[t]
    \centering
    \includegraphics[width=\linewidth]{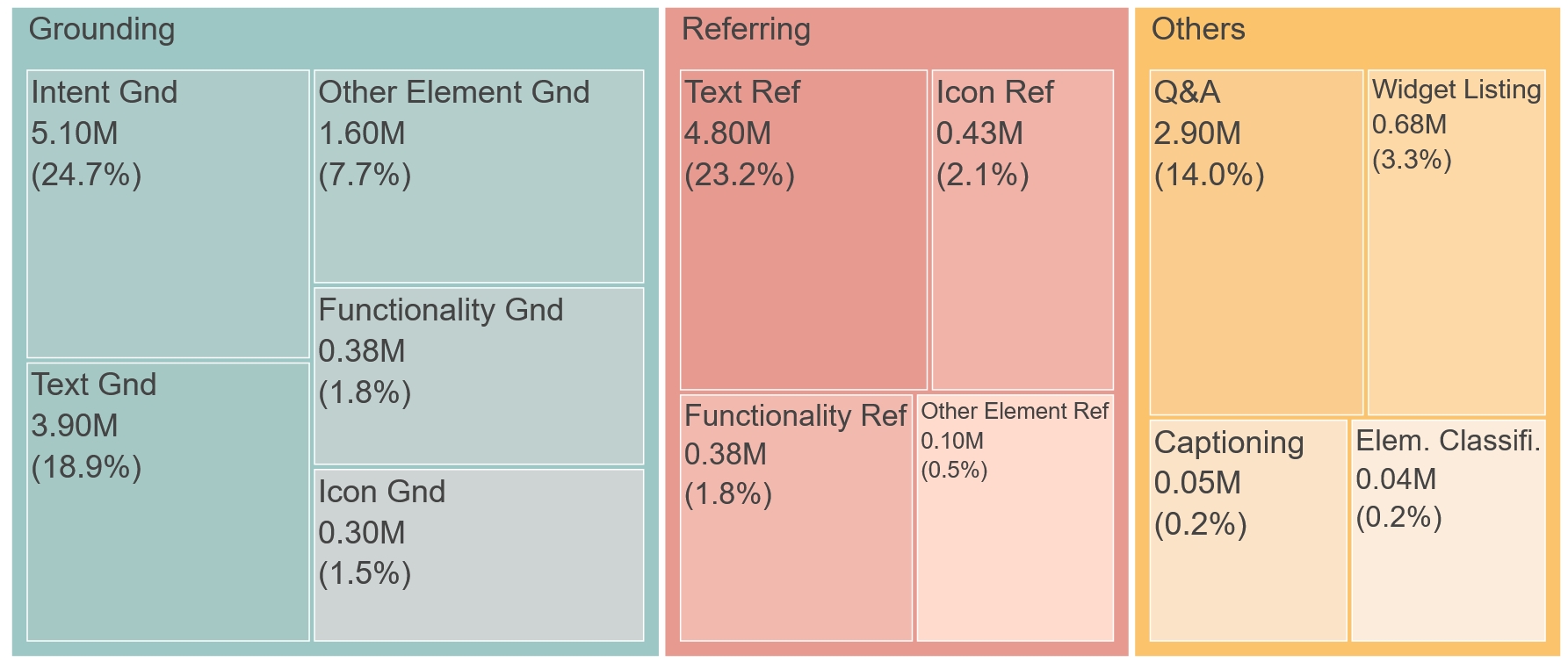}
    \caption{The proportions of the task types in our curated 20.6M-sample GUI understanding dataset introduced in Sec.~\ref{sec:gnd data construc}. Please refer to Sec.~\ref{sec:supp: data details} in Appendix for data sources and detailed statistics.}
    \label{fig: data treemap}
\end{figure}

\textbf{Data Overview} We ultimately contribute \textbf{20.6M} GUI understanding task samples associated with 3.3M clean elements from 2.5M unique GUI screenshots (see the statistics in Tab.~\ref{tab:data comparison}). The task proportions are shown in Fig.~\ref{fig: data treemap}.
Note that \textbf{67\% of the samples are newly annotated by the authors} and 33\% are collected and cleaned from existing open-source datasets.
For GUI agent task fine-tuning, we produce a 380k-size dataset integrating six sources for mobile tasks and a 145k-size one integrating three sources for web tasks. Data details are listed in Sec.~\ref{sec:supp: data details} and full dataset statistics are provided in Tables~\ref{tab:UI gnd data stats} and~\ref{tab:agent tasks} in Appendix.
\section{Experiments}

This section evaluates UIPro as an advanced GUI agent through extensive experiments: Sec.~\ref{sec:agent task eval} covers GUI agent tasks, Sec.~\ref{sec:gnd evaluation} examines GUI grounding, and Sec.~\ref{sec: ablation studies} provides ablation studies and analysis.

\subsection{Fine-Tuning Settings}
Since UIPro-SLiME begins as a tabula rasa VLM, we initially use LLaVA-Pretrain~\cite{liu2023llava} to train the VL-projector, then use RefCOCO~\cite{coco} data to fine-tune UIPro-SLiME with the visual encoder frozen, and finally continue fine-tuning it with our GUI understanding samples for one epoch.
For UIPro-Qwen2VL, as Qwen2-VL-7B~\cite{qwen2vl} already possesses strong visual understanding capabilities, we fine-tune it for one epoch with a 4.4M subset randomly extracted from the grounding tasks of the full set. The two base models are both fine-tuned with a learning rate of 3e-5. All coordinates are normalized in the range $[0,1000]$.

In the agent-task fine-tuning stage, we adhere to the training sample formatting outlined in SeeClick~\cite{cheng2024seeclick}, which includes the task and action history in the prompt and formats the ground truth action as a \verb|JSON| object recording the action type and arguments.
Action definitions are not included in the prompt since UIPro is fine-tuned and tested separately for web and mobile platforms. Our preliminary experiments indicated that excluding the action space leads to more efficient training. We fine-tune UIPro for six epochs until performance on downstream tasks plateaus. Additional hyperparameters are detailed in the Appendix.

\subsection{GUI Agent Task Evaluation}

\label{sec:agent task eval}
\begin{table}[]
\centering
\resizebox{\linewidth}{!}{%
\begin{tabular}{@{}cc|ccccc|c@{}}
\toprule
Methods     & Size  & General & Install & GoogleApps & Single & WebShop & Overall \\ \midrule

\rowcolor[HTML]{EFEFEF} 
GPT-4V-SoM~\cite{wonderland}      & -       & 41.7    & 42.6    & 49.8       & 72.8   & 45.7        & 50.5    \\
\rowcolor[HTML]{EFEFEF} 
GPT-4V-OmniParser~\cite{OmniParser} & - & 48.3 & 57.8 & 51.6 & 77.4 & 52.9 & 57.7 \\

Qwen2-VL~\cite{qwen2vl}     & 7B     & 22.0    & 27.9    & 24.6       & 32.7   & 18.2        & 25.1    \\
Fuyu-GUI~\cite{GUICourse} & 8B  &  -   & 50.9    & 41.6       & 45.7   & 43.8      & 45.5 \\
SeeClick~\cite{cheng2024seeclick}    & 10B     & 54.0    & 66.4    & 54.9       & 63.5   & 57.6        & 59.3    \\
OS-ATLAS~\cite{osatlas}    & 7B     & 57.9    & 63.4   & 55.5       & 79.1   & 59.7        & 63.1    \\

UIPro-Qwen2VL (ours)      & 7B    &  \textbf{64.4}   &  \textbf{74.6}   &   \textbf{67.9}     &   \textbf{79.4}  &  \textbf{67.6}      &   \textbf{70.4}  \\
\midrule 
MiniCPM-GUI~\cite{GUICourse} & 3B     & -    & 62.3    & 46.5       & 67.3   & 57.5       & 58.4    \\
UIPro-SLiME (ours)      & 3B    &  \textbf{67.0}   &  \textbf{71.4}   &   \textbf{65.4}     &   \textbf{73.2}  &  \textbf{62.9}      &   \textbf{68.0}  \\
\bottomrule
\end{tabular}
}

\caption{\textbf{Comparison on the AITW benchmark~\cite{rawles2023android}.} UIPro surpasses the compared methods by a large margin and even outperforms the strong GPT-4V-OmniParser~\cite{OmniParser}. The Step SR metric is reported for the five splits, and the Overall column denotes the Step SR calculated over all splits.}
\label{tab:AITW comp}
\end{table}
\vspace{-1mm}
\begin{table}[]
\centering
\resizebox{\linewidth}{!}{%
\begin{tabular}{@{}cc|ccc|ccc@{}}
\toprule
\multirow{2}{*}{Methods} & \multirow{2}{*}{Size} & \multicolumn{3}{c|}{High-Low} & \multicolumn{3}{c}{Only High} \\
 &  & Type $\uparrow$ & Click $\uparrow$ & Step SR $\uparrow$ & Type $\uparrow$ & Click $\uparrow$ & Step SR $\uparrow$ \\ \midrule

Qwen2-VL~\cite{qwen2vl} & 7B & 70.1 & 31.9 & 44.2 & 46.9 & 15.8 & 22.1 \\
OS-Atlas~\cite{osatlas} & 7B & 83.7 & 59.0 & 62.4 & 83.7 & \textbf{59.0} & 62.4 \\
UIPro-Qwen2VL (ours) & 7B & \textbf{96.3} & \textbf{84.3} & \textbf{85.5} & \textbf{83.8} & 57.2 & \textbf{64.0} \\ \midrule

Qwen2-VL~\cite{qwen2vl} & 2B & 34.9 & 9.8 & 11.3 & 22.0 & 4.1 & 6.1 \\
ShowUI~\cite{showui} & 2B & 47.2 & 43.3 & 29.5 & 45.3 & 33.0 & 22.9 \\
UIPro-SLiME (ours) & 3B & \textbf{98.1} & \textbf{44.1} & \textbf{61.1} & \textbf{84.6} & \textbf{34.5} & \textbf{46.0} \\ \bottomrule
\end{tabular}%
}

\caption{\textbf{Comparison on AndroidControl~\cite{androidcontrol}}. UIPro outperforms the other methods in step SR on the two settings.
Type and Click denote the accuracy (\%) of predicting action types and predicting the two actions requiring element localization (i.e., click and long-press), respectively. Evaluation is repeated three times, and the average is reported.}
\label{tab:andcon comp}
\end{table}
\vspace{-1mm}
\begin{table}[]
\centering
\resizebox{\linewidth}{!}{%
\begin{tabular}{@{}cc|ccc|ccc@{}}
\toprule
\multirow{2}{*}{Methods} & \multirow{2}{*}{Size} & \multicolumn{3}{c|}{GUIAct-Web} & \multicolumn{3}{c}{GUIAct-Mobile} \\
 &  & Type $\uparrow$ & Ground $\uparrow$& Step SR $\uparrow$& Type $\uparrow$& Ground $\uparrow$& Step SR $\uparrow$ \\ \midrule
GPT-4o & - & 77.1 & 45.0 & 41.8 & 63.0 & 58.2 & 44.3 \\

Qwen2-VL~\cite{qwen2vl} & 7B & 63.7 & 27.9 & 23.6 & 51.2 & 21.5 & 19.7 \\

UIPro-Qwen2VL (ours) & 7B & \textbf{85.1} & \textbf{82.7} & \textbf{69.1} & \textbf{84.2} & \textbf{78.4} & \textbf{67.2} \\ 

\midrule
ShowUI~\cite{showui} & 2B & 69.4 & 61.1 & 48.8 & 65.4 & 50.3 & 36.5 \\

MiniCPM-GUI~\cite{GUICourse} & 3B & 79.4 & 59.1 & 60.2 & 71.7 & 53.3 & 44.7 \\

UIPro-SLiME (ours) & 3B & \textbf{97.8} & \textbf{70.1} & \textbf{68.2} & \textbf{81.2} & \textbf{75.3} & \textbf{65.2} \\ \bottomrule
\end{tabular}%
}
\vspace{-2mm}
\caption{\textbf{Comparison on GUIAct~\cite{GUICourse}}. UIPro achieves notably higher step SR compared to the methods with equal model sizes.}
\label{tab:guicourse comp}
\end{table}
\begin{table*}[]
\centering

\resizebox{\textwidth}{!}{%
\begin{tabular}{@{}ccccccccccc@{}}
\toprule
\multirow{2}{*}{Methods} & \multirow{2}{*}{Size}  & \multicolumn{3}{c}{Cross-Task} & \multicolumn{3}{c}{Cross-Website} & \multicolumn{3}{c}{Cross-Domain} \\ \cmidrule(l){3-11} 
 &  & Elem. Acc.$\uparrow$ & Op. F1$\uparrow$ & Step SR$\uparrow$ & Elem. Acc.$\uparrow$ & Op. F1$\uparrow$ & Step SR$\uparrow$ & Elem. Acc.$\uparrow$ & Op. F1$\uparrow$ & Step SR$\uparrow$ \\ \midrule
\rowcolor[HTML]{EFEFEF} 
GPT-4V + SoM~\cite{OmniParser}  & - & - & - & 32.7 & - & - & 23.7 & - & - & 20.3 \\
\rowcolor[HTML]{EFEFEF} 
OmniParser (GPT-4V)~\cite{OmniParser}  & - &  42.4 & 87.6 & 39.4 & 41.0 & 84.8 & 36.5 & 45.5 & 85.7 & 42.0 \\
\rowcolor[HTML]{EFEFEF} 
UGround (GPT-4o)~\cite{uground}  & -  & 47.7 & - & - & 46.0 & - & - & 46.6 & - & - \\

Qwen-VL~\cite{bai2023qwen}  & 10B & 15.9 & 86.7 & 13.3 & 13.2 & 83.5 & 9.2 & 14.1 & 84.3 & 12.0 \\
SeeClick~\cite{cheng2024seeclick}  & 10B   & 28.3 & 87.0 & 25.5 & 21.4 & 80.6 & 16.4 & 23.2 & 84.8 & 20.8 \\
Fuyu-GUI~\cite{GUICourse}  & 8B   & 19.1 & 86.1 & 15.6 & 13.9 & 80.7 & 12.2 & 14.2 & 83.1 & 11.7 \\
UIX-Qwen2VL~\cite{multiUI} & 7B & 43.4 & - & 38.2 & 39.2 & - & 31.0 & 40.4 & - & 34.9 \\
OS-ATLAS~\cite{osatlas} & 7B & 39.7 & 80.0 & 36.7 & 39.5 & 78.1 & 35.7 & 40.4 & 79.5 & 37.2 \\
UIPro-Qwen2VL (ours)  & 7B  & \textbf{52.1} & \textbf{89.4} & \textbf{48.4} & \textbf{47.8} & \textbf{85.5} & \textbf{43.6} & \textbf{51.5} & \textbf{86.7} & \textbf{45.5} \\ \midrule
MiniCPM-GUI~\cite{GUICourse}  & 3B & 23.8 & 86.8 & 20.8 & 20.3 & 81.7 & 17.3 & 17.9 & 74.5 & 14.6 \\
UIPro-SLiME (ours) & 3B & \textbf{31.8} & \textbf{87.2} & \textbf{28.7} & \textbf{25.0} & \textbf{82.9} & \textbf{20.7} & \textbf{24.7} & \textbf{84.6} & \textbf{22.0} \\ \bottomrule
\end{tabular}%
}
\vspace{-2mm}
\caption{\textbf{Comparison on Multimodal-Mind2Web~\cite{deng2024mind2web}.} UIPro-Qwen2VL-7B outperforms existing methods of equal and larger size and surpasses the methods that use proprietary models. Elem. Acc. denotes the percentage of samples for which the models predict correct target coordinates. Op. F1 denotes the token-level F1 score for the predicted action.}
\label{tab:mind2web comp}
\end{table*}

\subsubsection{Benchmarks And Compared Methods}

We evaluate on three \textbf{mobile device control} benchmarks:
\begin{itemize}
    \item \textbf{AITW}~\cite{rawles2023android}: This large Android control dataset covers multiple scenarios, including App Store operations, browser use, and web shopping, across more than 350 apps. We use the same train/test split as in SeeClick~\cite{cheng2024seeclick}.
    \item \textbf{AndroidControl}~\cite{androidcontrol}: This Android dataset comprises 15,000 unique tasks over 833 apps, recorded from human demonstrations for training and evaluation. In addition to high-level tasks, AndroidControl provides low-level instructions for each step. We test on two settings: one with only high-level tasks and one with both high-level tasks and low-level instructions. We also follow AndroidControl~\cite{androidcontrol} to use a random-500 split for testing. Three runs are conducted, and the average performances are reported.
    \item \textbf{GUIAct-Smartphone}~\cite{GUICourse} also requires agents to complete mobile app operation tasks. It is generated by cleaning and re-annotating a subset of the AITW dataset~\cite{rawles2023android}.
\end{itemize}

These \textbf{web browser control} benchmarks are used:

\begin{itemize}
    \item \textbf{GUIAct-Web}~\cite{GUICourse}: This dataset collects websites, uses large language models (LLMs) to produce information-searching tasks, and employs annotators to label ground truth actions for evaluation.
    \item \textbf{Multimodal-Mind2Web}~\cite{deng2024mind2web} provides ground truth trajectories from human demonstrations for offline evaluation. It tests GUI agents on three test splits: unseen tasks, websites, and domains.
\end{itemize}

These benchmarks assess agents' ability to predict the next actions given the current GUI screenshot image, user tasks, and action history. The major evaluation metric is \textbf{Step Success Rate (Step SR)}~\cite{deng2024mind2web,rawles2023android}, where a generated step is considered correct only if the action type and arguments match ground truths. Step SR is calculated as the ratio (\%) of successful steps against all steps. More implementations are provided in the Appendix.

For \textbf{AITW},  these methods are compared: a) GPT-4V-SoM~\cite{wonderland} that detects GUI elements with IconNet~\cite{iconnet} and then performs Set-of-Marks prompting~\cite{setofmark}. b) OmniParser~\cite{OmniParser} that employs a GUI parser to detect interactable regions to enhance SoM prompting. c) GUI-oriented VLMs, such as SeeClick~\cite{cheng2024seeclick}, Fuyu-GUI~\cite{GUICourse}, and OS-ATLAS~\cite{osatlas}. For \textbf{AndroidControl}, these methods are compared: a) OS-ATLAS~\cite{osatlas} fine-tuned based on Intern-VL~\cite{chen2023internvl}. b) Qwen2-VL~\cite{qwen2vl}: a general VLM trained with GUI agent capability. c) ShowUI~\cite{showui} developed on Qwen2-VL-2B~\cite{qwen2vl}. For \textbf{GUIAct}, MiniCPM-GUI~\cite{GUICourse}, ShowUI~\cite{showui}, and Qwen2-VL~\cite{qwen2vl} are compared. Likewise, for \textbf{Mind2Web}, we also compare with methods equipped with proprietary models and expert VLMs dedicated to GUI agent tasks. As OS-ATLAS~\cite{osatlas} has not been evaluated on Mind2Web, we test this model by adding the Mind2Web action space to its system prompt.

\begin{table*}[t]
\small
\centering
\resizebox{\linewidth}{!}{%
\begin{tabular}{@{}ccc|ccccccc@{}}
\toprule
Model       & Size        & Input Res. & FuncGnd & ScreenSpot & ScreenSpot-v2  & MOTIF & RefExp & VWB EG   & VWB AG \\ \midrule
GPT-4o & - & AnyRes & 9.8 & 17.8 & 20.4 & 30.5 & 21.8 & 5.6 & 6.8 \\
Qwen2VL~\citep{qwen2vl} & 72B      & AnyRes           & 47.7     & 71.4 & 73.2 & 80.3 & 77.7 & 60.5 & 62.1  \\ 
Qwen2VL~\citep{qwen2vl} & 7B      & AnyRes           & 38.7  & 66.4 & 66.9 &75.1 & 64.8 & 55.9 & 62.1     \\ 

CogAgent~\citep{hong2023cogagent} & 18B     & 1120             &  29.3     & 47.4   & 49.2  &  46.7  & 35.0   &  55.7     & 59.2   \\
SeeClick~\citep{cheng2024seeclick} & 10B       & 448              & 19.8     &  53.4  &  54.0  & 11.1  &  58.1   & 39.2     & 27.2   \\
Ferret-UI~\cite{you2024ferretui} & 8B    & AnyRes           &  1.2  &  7.1 & 7.8 & 15.9  &  5.5 &  3.9  & 1.9    \\
UGround~\cite{uground} & 7B & AnyRes           &  48.8  &  74.8  & 76.5    & 72.4 &  73.6  & 85.2 &  63.1   \\

OS-ATLAS-Base~\cite{osatlas} & 7B & AnyRes           &  52.1  &  82.5   & 84.1    & 78.8 &  66.5  & 82.6 &  69.9   \\

UIPro-Qwen2VL (ours) & 7B & AnyRes & \textbf{58.8} & \textbf{82.5} & \textbf{86.9} & \textbf{80.6} & \textbf{81.9} & \textbf{94.9} & \textbf{70.9} \\ \midrule

Qwen2-VL~\cite{you2024ferretui} & 2B    & AnyRes           &  7.1   &  17.9  & 18.6 &  28.8 &  29.2  & 17.9       & 17.5   \\

UIPro-SLiME (ours) & 3B    & AnyRes           &    \textbf{58.3}   &  \textbf{60.7}    & \textbf{61.1}  &  \textbf{73.3}  &  \textbf{59.0}   &   \textbf{60.0}     & \textbf{40.8}   \\

\bottomrule
\end{tabular}
}
\caption{\textbf{Comparison on the GUI element grounding benchmarks.} UIPro achieves impressive grounding accuracy, especially on FuncPred, RefExp, and VWB EG. AnyRes means using an image division strategy to handle images with variable resolutions.}
\label{tab:gnd comparison}

\end{table*}

\subsubsection{Experimental Results}

Tables~\ref{tab:AITW comp}, ~\ref{tab:andcon comp}, and~\ref{tab:guicourse comp} show that UIPro achieves superior step SR on the mobile benchmarks.
On AITW, UIPro-Qwen2VL leads in performance, surpassing the GPT-4V-OmniParser~\cite{OmniParser}, which employs Omniparser to detect elements as a Set-of-Marks for GPT-4V.
Additionally, UIPro-Qwen2VL exceeds OS-ATLAS based on Qwen2-VL-7B by 7.7 in overall step SR.
On AndroidControl and GUIAct-Mobile, UIPro maintains high step SR, outperforming methods using proprietary models and GUI-oriented VLMs.

Tables~\ref{tab:guicourse comp} and~\ref{tab:mind2web comp} highlight UIPro's strong action prediction capabilities in web scenarios. UIPro surpasses competitors assisted by proprietary VLMs (in gray) and outperforms GUI-oriented VLMs of equal or larger sizes. The improvements on the Mind2Web splits over previous models, including OS-ATLAS and MiniCPM-GUI, underscore UIPro’s ability to generalize to unseen web interfaces.

\subsubsection{Error Analysis}
Analyzing the correctness of every action type, we found three significant error patterns: (a) Almost hit the target. The predicted point is outside but close to the ground truth bounding box. (b) Difficulty in using long-tailed actions, such as drag and hotkey, mainly due to insufficient training data. (c) The benchmarks, especially AITW~\cite{rawles2023android}, often fail to consider alternative solutions, leading to slightly inaccurate evaluations. 

\subsection{GUI Grounding Evaluation}
\label{sec:gnd evaluation}
We assess UIPro's GUI grounding capability after fine-tuning with the GUI understanding data in Sec.~\ref{sec:gnd data construc}, using six benchmarks:
\textbf{ScreenSpot}~\citep{cheng2024seeclick} and \textbf{ScreenSpot-v2}~\cite{osatlas} involve mobile, desktop, and web scenarios, requiring models to locate elements based on brief descriptions.
\textbf{MOTIF}~\citep{Burns2022ADF} prompts models to locate targets based on language intents in mobile apps.
\textbf{RefExp}~\citep{Bai2021UIBertLG} locates elements on mobile devices given action intents.
\textbf{VisualWebBench}~\citep{liu2024visualwebbench} provides element and action grounding tasks in web environments.
\textbf{FuncPred}~\cite{li2025autoguiscalingguigrounding} features challenging tasks requiring models to locate elements specified by functionality descriptions. Task examples are visualized in Fig.~\ref{fig:gnd example} of the Appendix.

We report grounding accuracy (\%): $\text { Acc }=\sum_{i=1}^N \mathbf{1}\left(\text {pred}_i \text { inside GT } \text { bbox }_i\right) / N \times 100$
where $\mathbf{1}$ is an indicator function and $N$ the number of test samples. This formula calculates the percentage of samples for which the predicted points fall within the elements' bounding boxes.

Tab.~\ref{tab:gnd comparison} shows that UIPro demonstrates higher accuracy on the GUI grounding benchmarks compared to competing models
UIPro-Qwen2VL outperforms the previous leading model, OS-ATLAS~\cite{osatlas}, with improvements of \textbf{15.4} and \textbf{12.3} on RefExp and VWB EG, respectively.
Notably, OS-ATLAS is fine-tuned with 13.8M element annotations, significantly more than the 4.4M used for UIPro-Qwen2VL.
Despite being only one-fifth the size, UIPro-SLiME surpasses CogAgent~\cite{hong2023cogagent} across benchmarks except for VWB AG.
Moreover, UIPro outperforms the other models on FuncPred~\cite{autoui}, benefiting from the incorporation of more functionality grounding tasks in the training data (see Sec.~\ref{sec:gnd data construc}). UIPro also achieves higher accuracy on the comprehensive ScreenSpot benchmark, with impressive performance on icon grounding tasks, shown in the Tab.~\ref{tab:sspot detail}.

Overall, the results demonstrate UIPro’s strong grounding capability across mobile and web platforms.

\subsection{Ablations And Analysis}
\label{sec: ablation studies}

This subsection examines the effectiveness of the 20.6M GUI understanding data (Sec.~\ref{sec:gnd data construc}), the unified action space (Sec.~\ref{sec:unify agent data}), scaling effects, and the denoising approach (Sec.~\ref{sec: denoise}). UIPro-SLiME is used for ablation unless otherwise stated.

\begin{figure}[t]
    \centering
    \includegraphics[width=0.9\linewidth]{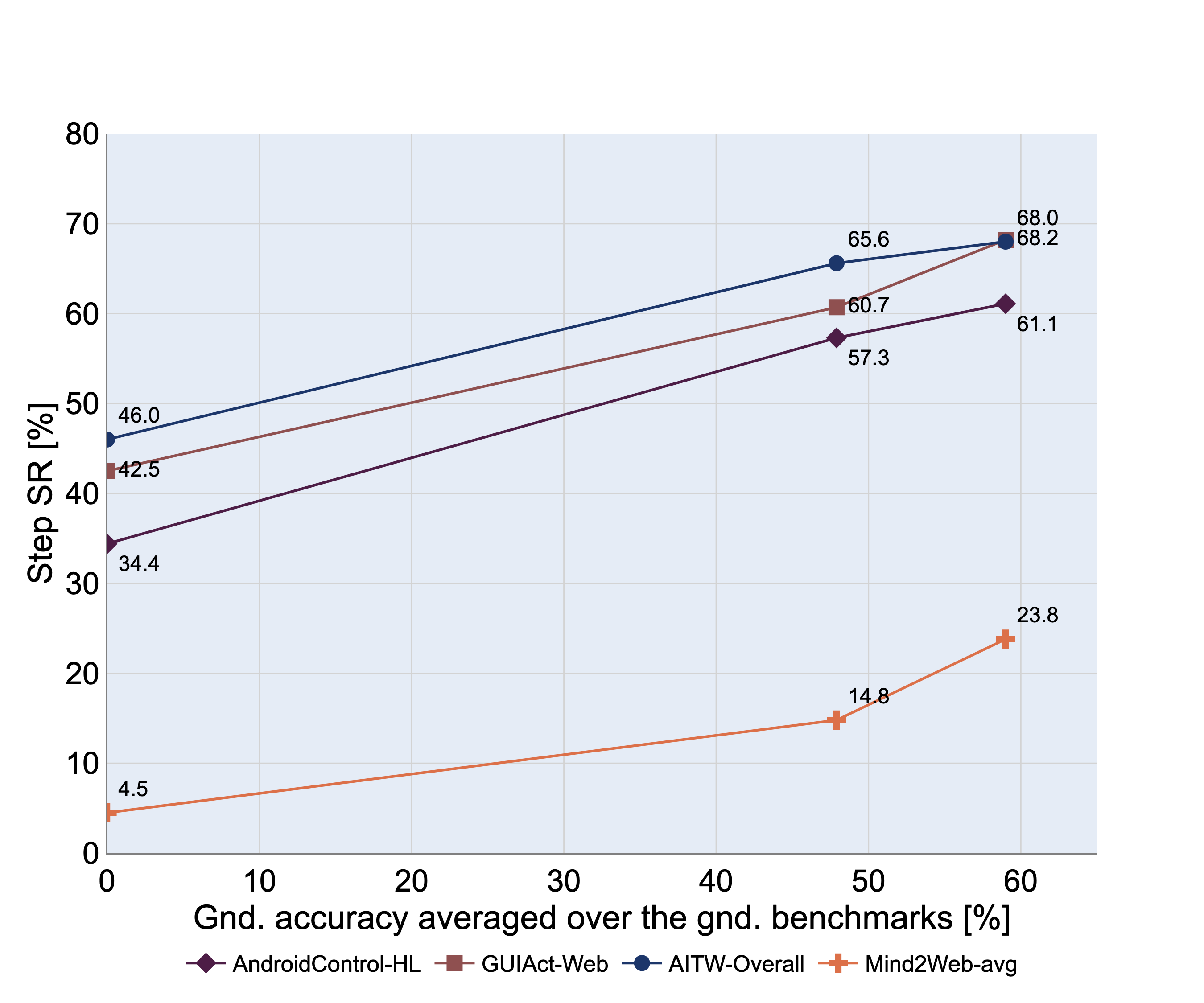}
    \vspace{-1mm}
    \caption{\textbf{Correlation between agent task performance and grounding accuracy.} UIPro-SLiME is pre-trained with 0, 5.9M, and 20.6M GUI understanding data, respectively, and then fine-tuned on the GUI agent tasks. We can see that higher grounding accuracy after pre-training leads to higher step SR after fine-tuning.}
    \label{fig: plan_vs_gnd}
\end{figure}

\noindent \textbf{Impact of GUI Understanding Pre-Training}
To demonstrate the benefits of the collected 20.6M GUI understanding data for downstream GUI agent task fine-tuning, we compare UIPro with two variants: one pre-trained without any GUI understanding data and another with only a 5.9M subset.
The results in Fig.~\ref{fig: plan_vs_gnd} show that without pre-training on GUI understanding data, UIPro exhibits significantly lower Step SR across all benchmarks. Increasing the pre-training data amount not only improves the average GUI grounding accuracy but also increases the downstream agent task performance. These findings suggest that UIPro achieves stronger task performance when fine-tuned on a foundation with higher grounding accuracy, consistent with SeeClick~\cite{cheng2024seeclick}.

\noindent \textbf{Ablations of Unified Action Space}
To assess the impact of the unified action space during GUI agent task fine-tuning, we mix the GUI agent task sources without unifying the action spaces for fine-tuning. Specifically, we retain the action definitions, including names and arguments, for each data source. We also compare a variant that uses only the training set of each GUI agent benchmark for fine-tuning.

\begin{figure}[t]
    \centering
    \includegraphics[width=0.95\linewidth]{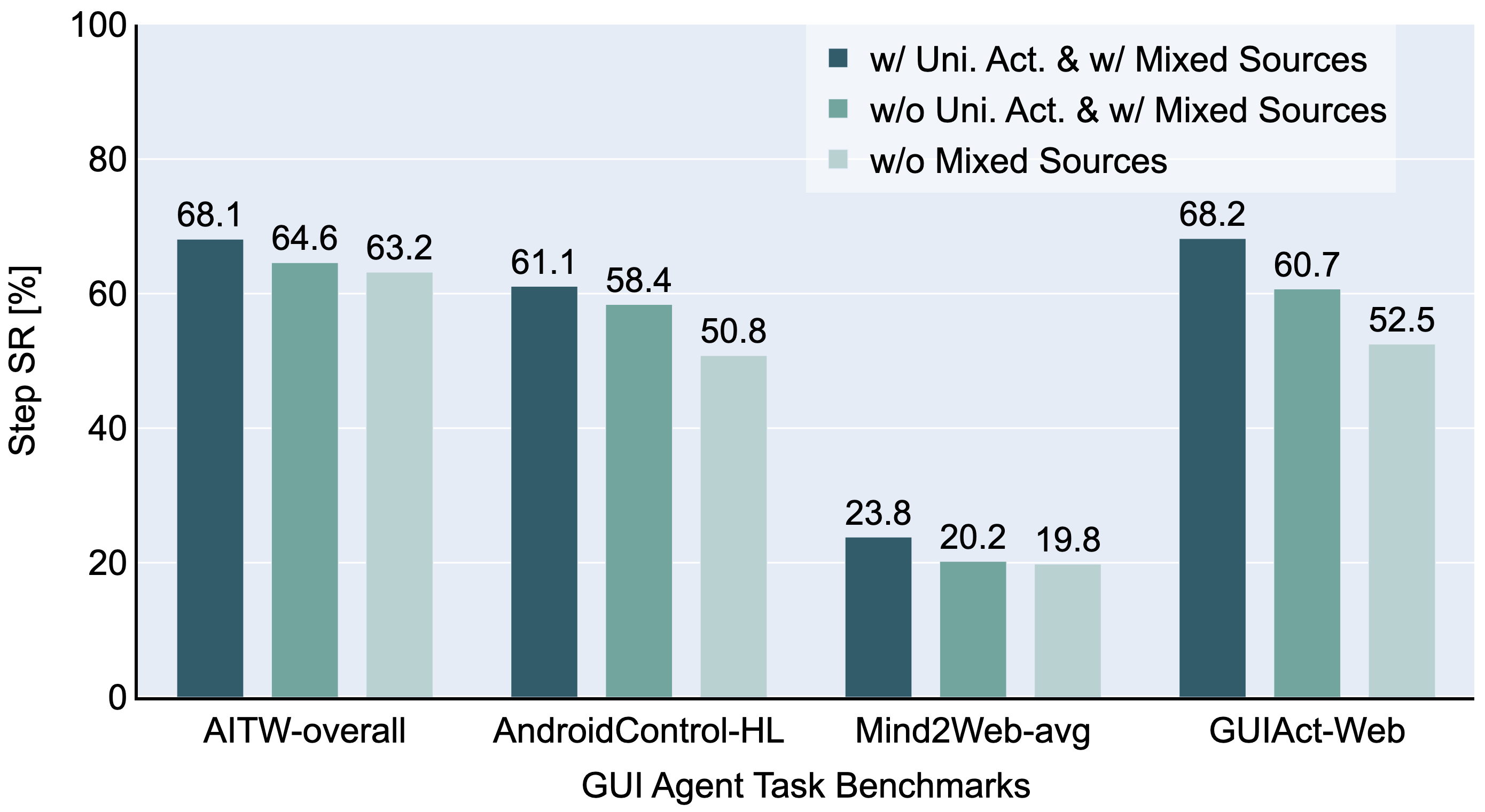}
    \caption{\textbf{Ablation study of the unified action spaces.} Fine-tuning UIPro with solely the training split of each benchmark (w/o Mixed Sources) is significantly inferior to the full UIPro. When mixing the data sources without unifying the heterogenous action spaces, UIPro witnesses clear decreases on all the benchmarks.}
    \label{fig: ablate uniact}
\end{figure}

The results in Fig.~\ref{fig: ablate uniact} show that mixing the different data sources achieves higher step SR than fine-tuning solely with the benchmark's training split. Without unifying the action spaces of the data sources, UIPro experiences notable performance declines across the benchmarks. Analysis of cases where UIPro without unified action space failed, while the full UIPro succeeded, revealed two major causes of the decrease: 1) significantly lower action type accuracy due to action definition conflicts; 2) lower accuracy of predicting swiping directions due to inconsistent swipe usage. These results suggest that unifying the action space better unleashes the potential of diverse GUI agent task data sources.

\noindent \textbf{Why Unified Action Space Works.}
To understand how the unified action space improves performance on agent tasks, we analyze UIPro-7B's improvements on AndroidControl~\cite{androidcontrol} by distinguishing between common and uncommon actions.
Table~\ref{tab:rebuttal:abalte uni act} demonstrates that fine-tuning with unified action space data consistently outperforms training exclusively on the benchmark's training set. Crucially, accuracy increases even for the \textit{\textbf{Wait}} action—which is unique to AndroidControl—suggesting benefits extend beyond common actions. We attribute this improvement to two mechanisms: (1) cross-task knowledge transfer, where exposure to diverse GUI transitions improves the model's ability to identify appropriate waiting contexts; and (2) training regularization from heterogeneous data sources, which enhances model robustness.
These findings indicate that action unification mitigates rather than exacerbates overfitting to common actions.

\begin{table}[t]
\centering
\resizebox{\columnwidth}{!}{%
\begin{tabular}{@{}c|cccc|ccc@{}}
\toprule
                              & \multicolumn{4}{c|}{Android Control - High} & \multicolumn{3}{c}{GUIAct} \\ \cmidrule(l){2-8} 
\multirow{-2}{*}{Variant} &
  \cellcolor[HTML]{FFCCC9}Click &
  \cellcolor[HTML]{FFCCC9}Swipe &
  \cellcolor[HTML]{CEF8F7}{\color[HTML]{000000} Navigate Back} &
  \cellcolor[HTML]{CEF8F7}{\color[HTML]{000000} Wait} &
  \cellcolor[HTML]{FFCCC9}Click &
  \cellcolor[HTML]{FFCCC9}Swipe &
  \cellcolor[HTML]{CEF8F7}Answer \\ \midrule
UIPro w/ benchmark training set  & 45.1         & 22.6        & 33.2        & 43.6        & 42.8    & 40.2    & 12.6   \\
UIPro w/ Uni. Act. Space (full) & \textbf{57.2}         & \textbf{50.9}        & \textbf{50.7}        & \textbf{73.4}        & \textbf{48.1}    & \textbf{57.0}    & \textbf{43.0}   \\ \bottomrule
\end{tabular}%
}
\caption{Compare UIPro-7B fine-tuned with only the AndroidControl training set and with our unified dataset. Unified action benefits \textcolor{red}{common} and \textcolor{cyan}{uncommon} actions. The metric is the prediction accuracy for each action type.}
\label{tab:rebuttal:abalte uni act}
\end{table}
\begin{figure}[t]
    \centering
    \includegraphics[width=1.0\linewidth]{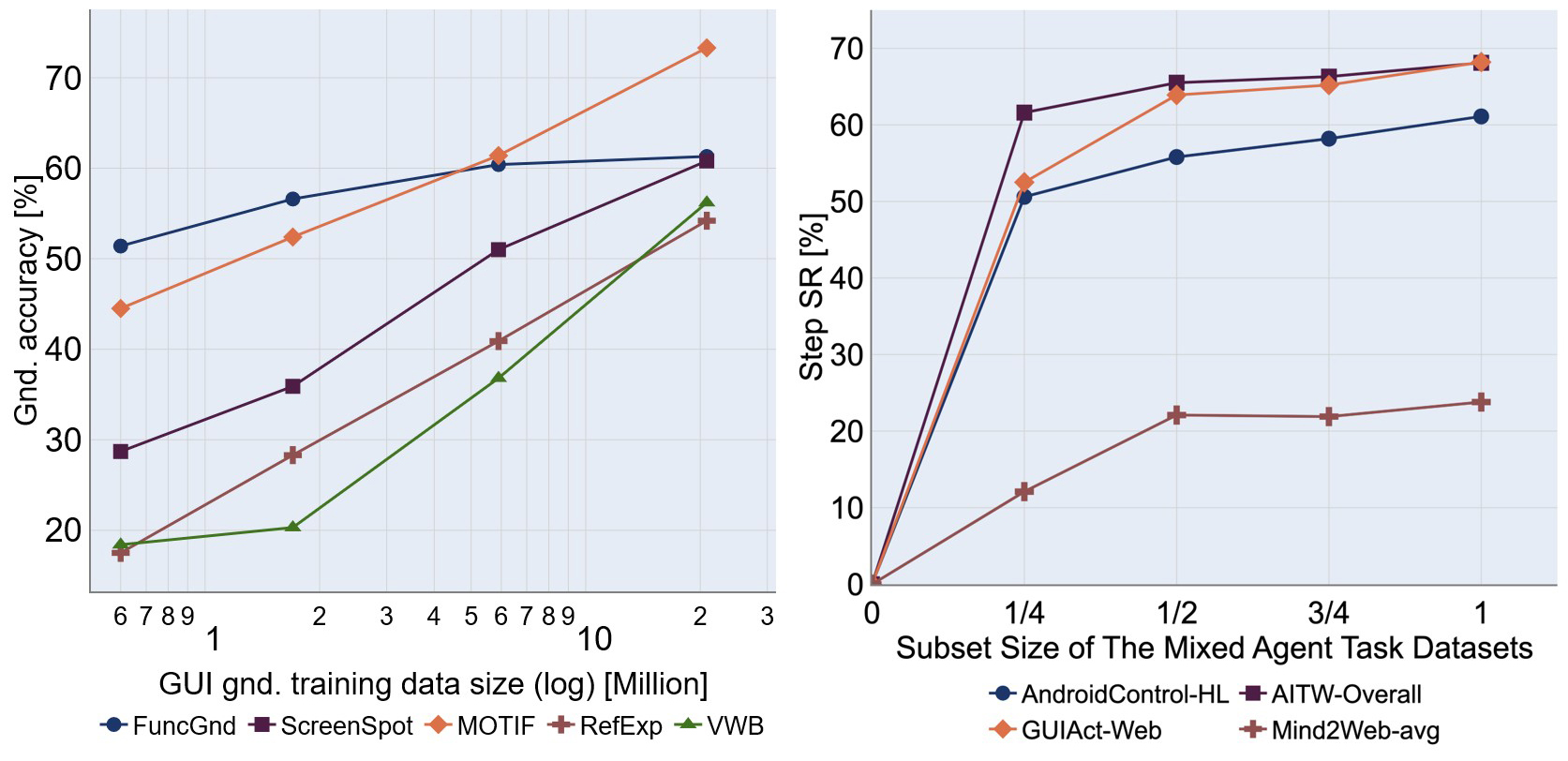}
    \vspace{-2mm}
    \caption{\textbf{Scaling effects of the GUI grounding data and mixed agent task data.} Left: increasing the data size of the GUI grounding data consistently improves grounding accuracy. Right: After pre-training on the 20.6M grounding data, fine-tuning UIPro with an increasing amount of the mixed GUI agent task data leads to higher step SR. Note that the step SR is zero as UIPro has not learned the output format required by the agent tasks.}
    \label{fig: scaling}
\end{figure}

\noindent\textbf{Scaling Effects of GUI Interaction Data.} \textbf{(a) On GUI Grounding.} To investigate how UIPro's grounding capability evolves during pre-training, we assess the training checkpoints at four data size checkpoints: 0.6M, 1.7M, 5.9M, and 20.6M. Fig.~\ref{fig: scaling} demonstrates that increasing the data size consistently improves the grounding accuracy. The scaling effect on the challenging FuncPred~\cite{autoui} is weaker as functionality annotations are more costly to acquire. \textbf{(b) On GUI Agent Task.} We also examine the scaling effects on mixed agent task data by fine-tuning UIPro-SLiME with $1/4$, $1/2$, $3/4$, and the full dataset. Fig.~\ref{fig: scaling} illustrates clear increasing trends on the four benchmarks.

\begin{table}[]
\scriptsize
\centering
\resizebox{\linewidth}{!}{%
\begin{tabular}{@{}ccccccc@{}}
\toprule
Variants           & FuncPred & ScreenSpot & MOTIF & RefExp & VWB EG & VWB AG \\ \midrule
w/ Denoise  & 66.7    & 56.4  & 69.2  & 45.8   & 52.3   & 34.0   \\
w/o Denoise & 64.3    & 55.7  & 66.9  & 43.5   & 50.6   & 32.0   \\ \bottomrule
\end{tabular}
}
\vspace{-2mm}
\caption{\textbf{Efficacy of the proposed denoising procedure.} UIPro-SLiME is pre-trained using 6.7M GUI understanding tasks with and without denoising. We can see that denoising contributes to notable grounding accuracy gains over all the benchmarks.}
\label{tab:ablate noise}
\end{table}
\noindent\textbf{Efficacy of Denoising} To validate the denoising procedure in Sec.~\ref{sec: denoise}, we pre-train UIPro with 6.7M GUI understanding tasks with and without denoising. Tab.~\ref{tab:ablate noise} highlights the importance and effectiveness of our denoising approach.

\section{Conclusion}
This paper introduces UIPro, a generalist GUI agent with superior GUI interaction capability. By curating extensive multi-platform and multi-task GUI interaction data and the proposed unified action space, UIPro exhibits superior performance on multiple GUI interaction and grounding benchmarks. We hope our curation programs and the cleaned dataset will facilitate further research and development in the GUI agent domain.

\section{Acknowledgments}
This work was supported in part by the National Key R\&D Program of China (No. 2022ZD0160102), the National Natural Science Foundation of China (No. U21B2042, No. 62320106010).
{
    \small
    \bibliographystyle{ieeenat_fullname}
    \bibliography{main}
}

\clearpage
\setcounter{page}{1}
\maketitlesupplementary

\section{Details of UIPro Datasets}
\label{sec:supp: data details}
We list the data sources used in the 20.6M GUI understanding dataset (Sec.~\ref{sec:gnd data construc}) in Tab.~\ref{tab:UI gnd data stats} and those in the unified GUI agent task (Sec.~\ref{sec:unify agent data}) in Tab.~\ref{tab:agent tasks}.

GUI understanding data:

\noindent\textbf{Common Crawl:} Following~\cite{li2025autoguiscalingguigrounding}, we select the pages from the top-200 domains in Common Crawl and design an in-house web crawler that interacts with elements rendered on the web page and collects interaction trajectories. Subsequently, we use the AutoGUI~\cite{li2025autoguiscalingguigrounding} pipeline to generate functionality grounding and referring tasks.

\noindent\textbf{Android Emulator:} We set up virtual Android phones on Android Emulator and collect interaction trajectories on GUIs, including the home page, drop-down panel, settings page, and Apps drawer. Likewise, we use the AutoGUI~\cite{li2025autoguiscalingguigrounding} pipeline to generate functionality grounding and referring tasks.

\noindent\textbf{RICO}~\cite{deka2017rico}: We use the element annotations prepared by SeeClick~\cite{cheng2024seeclick} and generate element grounding and referring tasks using RICO element descriptions as referring expressions.

\noindent\textbf{MobileViews}~\cite{mobileviews}: This dataset provides massive GUIs recorded on 20k mobile apps.We generate text localization, OCR, intent grounding, and widget listing tasks using the GUI metadata of this dataset.

\noindent\textbf{WAE}~\cite{chen2020wireframe}: This dataset also provides large-scale GUI metadata, which are originally used for assisting GUI design. Similar to MobileViews, we generate grounding and referring tasks from GUI metadata. As this data source provides accurate element properties. We also generate tasks specific to iconic elements.

\noindent\textbf{WebUI}~\cite{WebUI}: This dataset also provides large-scale GUI metadata. We generate GUI understanding tasks as we do for WAE.

\noindent\textbf{MultiUI}~\cite{multiUI}: This dataset provides massive GUI-related Q\&A tasks, which are cleaned and incorporated into our dataset to enhance the model's capability of understanding various aspects of GUIs.

\noindent\textbf{GUIEnv}~\cite{GUICourse}: This dataset contains only text localization and OCR tasks, which are both cleaned and incorporated into our dataset

\noindent\textbf{SeeClick-Web}~\cite{cheng2024seeclick}: This dataset contains only text localization and OCR tasks in web scenarios, which are cleaned and incorporated into our dataset.

\noindent\textbf{OmniAct}~\cite{kapoor2024omniact}: This dataset contains element annotations on web and desktop scenarios, which are used to generate intent grounding tasks.

\noindent\textbf{MOTIF}~\cite{Burns2022ADF}: This dataset contains action trajectories collected on mobile apps. We convert the actions into intent-grounding tasks.

GUI agent task data:

\noindent\textbf{AITW}~\cite{rawles2023android}: AITW is released by Google Research, providing massive Android app interaction trajectories. We use the train/test splits provided by SeeClick~\cite{cheng2024seeclick} and incorporate the cleaned training samples into our unified agent task data.

\noindent\textbf{AITZ}~\cite{AITZ}: This dataset cleans a subset of AITW~\cite{rawles2023android} and uses a proprietary LLM to generate high-quality reasoning and action annotations. Given a step in AITZ, We generate a sample with a reasoning process and one without reasoning to leverage the GUI knowledge entailed in the reasoning content.

\noindent\textbf{AMEX}~\cite{AMEX}: This dataset expands the General split of the AITW~\cite{rawles2023android} to provide more detailed annotations for each interaction step. The quality of this dataset is high, so we directly reformat their samples and incorporate them into our unified dataset.

\noindent\textbf{AndroidControl}~\cite{androidcontrol}: This dataset boasts massive, high-quality interaction trajectories collected over one year by Google Research. As this dataset has not provided the bounding box of target elements, we find the smallest box enclosing target points using the GUI metadata provided.

\noindent\textbf{GUIOdyssey}~\cite{GUIOdyssey}: This dataset provides cross-app interaction trajectories, which can diversify our unified dataset.

\noindent\textbf{WebLINX}~\cite{weblinx}: This dataset provides dialogue-format human-agent interaction trajectories. We remove all non-action steps and use the action-related steps to generate action prediction samples.

\noindent\textbf{OmniAct-Desktop}~\cite{kapoor2024omniact}: This dataset is used in the experiments to assess the transferability of UIPro. As each action plan provided in OmniAct-Desktop is associated with only the starting screenshot, we extract the first action in the plan to generate training and test data.

\section{Implementation Details of UIPro}
\subsection{Training Parameters}
\begin{table}[t]
\centering
\small
\caption{The training hyper-parameters used for fine-tuning UIPro-SLiME.}
\label{tab:training config slime}
\begin{tabular}{@{}cc@{}}
\toprule
Hyper-Parameter & Value \\ \midrule
Epoch & 1 \\
Global batch size & 128 \\
\#GPUs & 8 \\
Learning rate for all stages & 3e-5 \\
weight decay & 0.0 \\
ADAM Beta2 & 0.95 \\
Warm-up ratio & 0.03 \\
LR scheduler & Cosine \\
Model max length & 2048 \\
Frozen module & ViT \\
DeepSpeed & ZeRO-2 \\
Data type & BFloat16 \\ \bottomrule
\end{tabular}
\end{table}
\begin{table}[t]
\centering
\small
\caption{The training hyper-parameters used for fine-tuning UIPro-Qwen2VL.}
\label{tab:training config qwen2vl}
\begin{tabular}{@{}cc@{}}
\toprule
Hyper-Parameter & Value \\ \midrule
Epoch & 1 \\
Global batch size & 128 \\
\#GPUs & 8 \\
Learning rate for all stages & 3e-5 \\
LoRA Rank & 128 \\
LoRA Alpha & 256 \\
weight decay & 0.0 \\
ADAM Beta2 & 0.95 \\
Warm-up ratio & 0.03 \\
LR scheduler & Cosine \\
Model max length & 4096 \\
Frozen module & ViT \\
DeepSpeed & ZeRO-2 \\
Data type & BFloat16 \\ \bottomrule
\end{tabular}
\end{table}

The hyper-parameters of training UIPro-SliME and UIPro-Qwen2VL are shown in Tab.~\ref{tab:training config slime} and Tab.~\ref{tab:training config qwen2vl}. All experiments are conducted with 8 L20 GPUs, each with 48GB of memory. Pre-training UIPro with the 20.6M GUI understanding data for one epoch took approximately 96 hours on the 8 L20 GPUs; Fine-tuning UIPro with the 380k unified agent task data for the mobile embodiment took approximately 9 hours; Fine-tuning UIPro with the 144.9k unified agent task data for the web embodiment took approximately 3 hours.

\subsection{Unified Action Spaces}
\label{sec:supp: act space def}
We unify the heterogeneous action definitions from the used data sources and generate three unified action spaces for mobile, web, and desktop scenarios, respectively. Please refer to Tab.~\ref{tab: mobile act space}, Tab.~\ref{tab: web act space}, and Tab.~\ref{tab: desktop act space}.

Inconsistency mainly exhibits in swipe, scroll, drag/move, and status (used to signal task completion/impossibility), with substantially different parameter definitions on AITW~\cite{rawles2023android}, AndroidControl~\cite{androidcontrol}, GUIOdyssey~\cite{GUIOdyssey}, and Mind2Web~\cite{deng2024mind2web}.

\begin{table*}[]
\tiny
\centering
\resizebox{\textwidth}{!}{%
\begin{tabular}{@{}cll@{}}
\toprule
Action Name      & \multicolumn{1}{c}{Usage}                                                                                                                                                                                                                                                                               & \multicolumn{1}{c}{Definition}                                                                                                                                            \\ \midrule
click            & Click on an element. The target\_x and target\_y denote the x and y coordinates of the target.                                                                                                                                                                                                          & \{"action\_type": "click", "target": (\{target\_x\},\{target\_y\})\}                                                                                                      \\
\rowcolor[HTML]{EFEFEF} 
long\_press      & long-press an element for a duration.                                                                                                                                                                                                                                                                   & \{"action\_type": "long\_Press", "target": (\{target\_x\},\{target\_y\})\}                                                                                                \\
swipe            & \begin{tabular}[c]{@{}l@{}}Simulate a real swipe action to change viewports. The start\_x and start\_y denote the x and y\\ coordinates of the swipe starting point. The direction has four options: up, down, left, and right.\\ The distance has three options: short, medium, and long.\end{tabular} & \begin{tabular}[c]{@{}l@{}}\{"action\_type": "swipe", "start": (\{start\_x\},\{start\_y\}),\\     "direction": "\{direction\}", "distance": "\{distance\}"\}\end{tabular} \\
\rowcolor[HTML]{EFEFEF} 
input\_text      & Type texts in to an input box.                                                                                                                                                                                                                                                                          & \{"action\_type": "input\_text", "text": "\{text\}"\}                                                                                                                     \\
drag             & \begin{tabular}[c]{@{}l@{}}Press a finger on a target, move the finger to a destination,\\ and finally list the finger.\end{tabular}                                                                                                                                                                    & \{"action\_type": "drag", "start": (x1,y1), "end": (x2,y2)\}                                                                                                              \\
\rowcolor[HTML]{EFEFEF} 
enter            & \begin{tabular}[c]{@{}l@{}}This action inherits from AndroidControl and Android World.\\ It simulates pressing Enter on a keyboard.\end{tabular}                                                                                                                                                        & \{"action\_type": "enter"\}                                                                                                                                               \\
navigate\_back   & Navigate to the previous GUI.                                                                                                                                                                                                                                                                           & \{"action\_type": "navigate\_back"\}                                                                                                                                      \\
\rowcolor[HTML]{EFEFEF} 
navigate\_home   & Navigate to the home page on the mobile phone.                                                                                                                                                                                                                                                          & \{"action\_type": "navigate\_home"\}                                                                                                                                      \\
navigate\_recent & Open the window showing recently used apps.                                                                                                                                                                                                                                                             & \{"action\_type": "navigate\_recent"\}                                                                                                                                    \\
\rowcolor[HTML]{EFEFEF} 
wait             & Wait for content loading. This is also inherited from AndroidControl and Android World.                                                                                                                                                                                                                 & \{"action\_type": "wait"\}                                                                                                                                                \\
status           & \begin{tabular}[c]{@{}l@{}}Signal the termination of a task and return an answer if required.\\ The goal\_status has two options: "successful" denoting task completion and "infeasible"\\ denoting an impossible task. The answer is used to contain the answer generated by the model.\end{tabular}   & \{"action\_type": "status", "goal\_status": "\{goal\_status\}", "answer": "\{answer\}"\}                                                                                  \\ \bottomrule
\end{tabular}%
}
\caption{The unified action space for mobile scenarios.}
\label{tab: mobile act space}
\end{table*}
\begin{table*}[]
\tiny
\centering
\resizebox{\textwidth}{!}{%
\begin{tabular}{@{}cll@{}}
\toprule
Action Name       & \multicolumn{1}{c}{Usage}                                                                                                                                                                                   & \multicolumn{1}{c}{Definition}                                                           \\ \midrule
click             & Click on an element. The target\_x and target\_y denote the x and y coordinates of the target.                                                                                                              & \{"action\_type": "click", "target": (\{target\_x\},\{target\_y\})\}                     \\
\rowcolor[HTML]{EFEFEF} 
scroll            & \begin{tabular}[c]{@{}l@{}}Simulate a scroll action to change viewports. The direction has four options: up, down, left, and right.\\ The distance has three options: short, medium, and long.\end{tabular} & \{"action\_type": "scroll", "direction": "\{direction\}", "distance": "\{distance\}"\}   \\
input\_text       & Type texts into an input box.                                                                                                                                                                              & \{"action\_type": "input\_text", "text": "\{text\}"\}                                    \\
\rowcolor[HTML]{EFEFEF} 
drag              & \begin{tabular}[c]{@{}l@{}}Press a finger on a target, move the finger to a destination,\\ and finally lift the finger.\end{tabular}                                                                        & \{"action\_type": "drag", "start": (x1,y1), "end": (x2,y2)\}                             \\
move\_to          & This action simulates moving the mouse and can be used to move the pointer to a location or hover on an element.                                                                                            & \{"action\_type": "move\_to", "start": (x1,y1), "end": (x2,y2)\}                         \\
\rowcolor[HTML]{EFEFEF} 
navigate\_back    & Navigate to the previous webpage.                                                                                                                                                                           & \{"action\_type": "navigate\_back"\}                                                     \\
navigate\_forward & Undo navigate\_back.                                                                                                                                                                                        & \{"action\_type": "navigate\_home"\}                                                     \\
\rowcolor[HTML]{EFEFEF} 
go\_to            & Go to a certain URL.                                                                                                                                                                                        & \{"action\_type": "go\_to", "url": "(a certain url)"\}                                   \\
search\_google    & This action simulates directly typing a search query in the address bar and pressing Enter.                                                                                                                    & \{"action\_type": "search\_google", "query": "(search query)"\}                          \\
\rowcolor[HTML]{EFEFEF} 
press\_key        & \begin{tabular}[c]{@{}l@{}}This action simulates pressing a key down and then releasing it.\\ Example keys include 'enter', 'shift', arrow keys, or function keys.\end{tabular}                             & \{"action\_type": "press\_key", "key": "(key name)"\}                                    \\
hotkey            & Press a key combination. The key\_comb examples include Ctrl-S or Ctrl-Shift-1 with multiple keys combined with '-'.                                                                                        & \{"action\_type": "hotkey", "key\_comb": "(key combination)"\}                           \\
\rowcolor[HTML]{EFEFEF} 
new\_tab          & Create a new tab in the web browser                                                                                                                                                                         & \{"action\_type": "new\_tab"\}                                                           \\
switch\_tab       & Switch to a tab specified by its index.                                                                                                                                                                        & \{"action\_type": "switch\_tab", "tab": "(tab index)"\}                                  \\
\rowcolor[HTML]{EFEFEF} 
close\_tab        & Close the focused tab.                                                                                                                                                                                      & \{"action\_type": "close\_tab"\}                                                         \\
status            & Signal the termination of a task and return an answer if required. The usage of its parameters is the same as mobile.                                                                                       & \{"action\_type": "status", "goal\_status": "\{goal\_status\}", "answer": "\{answer\}"\} \\ \bottomrule
\end{tabular}%
}
\caption{The unified action space for web scenarios.}
\label{tab: web act space}
\end{table*}
\begin{table*}[]
\tiny
\centering
\resizebox{\textwidth}{!}{%
\begin{tabular}{@{}cll@{}}
\toprule
Action Name   & \multicolumn{1}{c}{Usage}                                                                                                                                                                                   & \multicolumn{1}{c}{Definition}                                                           \\ \midrule
click         & Click on an element. The target\_x and target\_y denote the x and y coordinates of the target.                                                                                                              & \{"action\_type": "click", "target": (\{target\_x\},\{target\_y\})\}                     \\
\rowcolor[HTML]{EFEFEF} 
right\_click  & Right-click on an element.                                                                                                                                                                                  & \{"action\_type": "right\_click", "target": (\{target\_x\},\{target\_y\})\}              \\
double\_click & Double-click on an element.                                                                                                                                                                                 & \{"action\_type": "double\_click", "target": (\{target\_x\},\{target\_y\})\}             \\
\rowcolor[HTML]{EFEFEF} 
scroll        & \begin{tabular}[c]{@{}l@{}}Simulate a scroll action to change viewports. The direction has four options: up, down, left, and right.\\ The distance has three options: short, medium, and long.\end{tabular} & \{"action\_type": "scroll", "direction": "\{direction\}", "distance": "\{distance\}"\}   \\
input\_text   & Type texts in to an input box.                                                                                                                                                                              & \{"action\_type": "input\_text", "text": "\{text\}"\}                                    \\
\rowcolor[HTML]{EFEFEF} 
drag          & \begin{tabular}[c]{@{}l@{}}Press a finger on a target, move the finger to a destination,\\ and finnaly list the finger.\end{tabular}                                                                        & \{"action\_type": "drag", "start": (x1,y1), "end": (x2,y2)\}                             \\
move\_to      & This action simulates moving the mouse and can be used to move the pointer to a location or hover on an element.                                                                                            & \{"action\_type": "move\_to", "start": (x1,y1), "end": (x2,y2)\}                         \\
\rowcolor[HTML]{EFEFEF} 
press\_key    & \begin{tabular}[c]{@{}l@{}}This action simulates pressing a key down and then releasing it.\\ Example keys include 'enter', 'shift', arrow keys, or function keys.\end{tabular}                             & \{"action\_type": "press\_key", "key": "(key name)"\}                                    \\
hotkey        & Press a key combination. The key\_comb examples include Ctrl-S or Ctrl-Shift-1 with multiple keys combined with '-'.                                                                                        & \{"action\_type": "hotkey", "key\_comb": "(key combination)"\}                           \\
\rowcolor[HTML]{EFEFEF} 
status        & Signal the termination of a task and return an answer if required. The usage of its parameters is the same as mobile.                                                                                       & \{"action\_type": "status", "goal\_status": "\{goal\_status\}", "answer": "\{answer\}"\} \\ \bottomrule
\end{tabular}%
}
\caption{The unified action space for desktop scenarios.}
\label{tab: desktop act space}
\end{table*}

\subsection{Denoising Procedure}
\label{sec:supp:denoise pipeline}
We inspect the used data sources listed in Tab.~\ref{tab:UI gnd data stats} and Tab.~\ref{tab:agent tasks}, categorize typical noise in the raw GUI data, and develop the following denoising procedure:

1. \textbf{Checking the validity of element bounding box coordinates.} An element will be discarded if it is outside of the GUI screenshot or its area is zero.

2. \textbf{Removing oversized elements.} As the grounding and referring tasks in our dataset focus on elements that are leaf nodes in the GUI layout hierarchy, we also remove oversized elements that are likely parent nodes. Localizing or referring to these oversized elements probably leads to ambiguity, as the element center often falls in leaf node elements. We remove elements whose area ratios are greater than 0.65. We choose 0.65 as this threshold can achieve an empirically good tradeoff between retaining meaningful elements and removing as many noisy ones as possible.

3. \textbf{Removing extremely tiny elements.} According to Google Accessibility Help\footnote{https://support.google.com/accessibility/android/answer/7101858?hl=en}, an element should be large enough for reliable interaction, with a width and height of at least 48dp. Considering that low-density GUI screenshots probably exist in the used GUI data sources, the smallest size of an element is limited to $48 \times (60/160)=18 \text{pixels}$. We remove elements whose smaller size is less than this threshold.

4. \textbf{Removing blank elements.} It is a common case that the GUI rendering is incomplete due to massive content loading and rendering program bugs. An example is the Mind2Web dataset~\cite{deng2024mind2web}, which contains many incomplete web page HTML scripts. Generating GUI interaction samples from these blank elements is likely to confuse the trained models. To remove these blank elements, we calculate the color deviation using \verb|np.std| for the element region and remove elements whose color deviation is less than 5 (also an empirical value to achieve a good tradeoff between retaining meaningful elements and removing noisy ones as many as possible).

5. \textbf{Removing duplicate boxes.} It is also a common case that multiple elements are in the same bounding box. For example, an image button may contain an image element and a button element. We retain only one of the duplicate elements to ensure the dataset's diversity.

6. \textbf{Removing invisible textual elements.} Invisible elements, e.g., a hidden menu, will confuse the model. To remove these elements, we utilize \verb|pytesseract| (an efficient OCR toll) to detect the texts for textual elements and remove the elements for which the similarity score between the OCR outputs and their text properties is less than 22.

7. \textbf{Denoising for agent task data.} Apart from the noise in GUI grounding data, the used GUI agent task data also contains noise. We list the most occurring noise type in the GUI agent task dataset: 1) AITZ~\cite{AITZ}: the action mentioned in the agent's reasoning content does not match the actually taken action. 2) AITW~\cite{rawles2023android}: Apart from the noise recognized by other works~\cite{AITZ,AMEX}, we also find that in some cases, one action is repeated multiple times, leading to redundant training samples. 3) AndroidControl~\cite{AMEX}: No bounding box associated with the interacted element. 4) Mind2Web~\cite{deng2024mind2web}: blank interacted elements.

Our denoising procedure is designed for the data sources used. Nevertheless, one can integrate more rules and adjust the empirical thresholds to extend our procedure to more data sources.

The noise ratios for several data sources are listed in Tab.~\ref{tab:noise ratio}. We find that the noise ratios are unignorable, with 29.0\% of WAE~\cite{chen2020wireframe} elements being noisy. 

\section{Additional Experiments}
\begin{table}[]
\small
\centering

\begin{tabular}{@{}ccc@{}}
\toprule
Methods    & Size & Step SR\\ \midrule
\rowcolor[HTML]{EFEFEF} 
DetAct (GPT-4V)~\cite{kapoor2024omniact} & - &  17.0 \\
\rowcolor[HTML]{EFEFEF} 
UGround (GPT-4o)~\cite{uground}  & -   & 33.4     \\

OS-ATLAS-Base w/ SFT~\cite{osatlas}  & 7B   & 74.6     \\

UIPro-Qwen2VL  & 7B   & \textbf{77.9}   \\ \midrule
MiniCPM-GUI~\cite{GUICourse}     & 3B   & 11.3      \\
UIPro-SLiME w/o GUI und. PT     & 3B   & 15.4      \\
UIPro-SLiME w/ 5.9M GUI und. PT     & 3B   & 18.9      \\
UIPro-SLiME w/ 20.6M GUI und. PT     & 3B   & \textbf{25.1}      \\ \bottomrule
\end{tabular}

\caption{\textbf{Evaluating UIPro on the desktop software tasks of OmniAct~\cite{kapoor2024omniact}}. Although training samples from desktop domains are scarce in our collected data, the two UIPro models still perform well. Pre-training UIPro-SLiME with more of our GUI understanding data before fine-tuning it on the downstream OmniAct tasks leads to better step SR. OS-ATLAS-Base w/ SFT means OS-ATLAS-Base~\cite{osatlas} is finetuned with the OmniAct training split.}
\label{tab:omniact comp}
\end{table}
\subsection{Transfer to Desktop Environments}

\noindent\textbf{Transfer UIPro to OmniAct-Desktop Tasks} Although UIPro is primarily pre-trained with web and mobile data, we test whether this pre-training can facilitate agent task fine-tuning on out-of-distribution desktop environments, such as Windows and Linux. We fine-tune the pre-trained UIPro with the training split of OmniAct-desktop~\cite{kapoor2024omniact} and evaluate it on the test split. Tab.~\ref{tab:omniact comp} shows that increasing the pre-training data size consistently improves step SR, with UIPro even surpassing OS-ATLAS\cite{osatlas}, which is pre-trained with extensive desktop-domain data and fine-tuned with the same OmniAct data.

\subsection{Detailed Performance on ScreenSpot}
\label{sec:supp:sspot detail}
The grounding accuracy on the three platform splits of ScreenSpot~\cite{cheng2024seeclick} is shown in Tab.~\ref{tab:sspot detail}. The results show that UIPro obtains the highest grounding accuracy and excels at icon grounding.

\begin{table*}[]
\centering
\begin{tabular}{@{}cc|cc|cc|cc|c@{}}
\toprule
\multirow{2}{*}{Methods} & \multirow{2}{*}{Model Size} & \multicolumn{2}{c|}{Mobile} & \multicolumn{2}{c|}{Desktop} & \multicolumn{2}{c|}{Web} & \multirow{2}{*}{Avg.} \\
                     &     & Text & Icon & Text & Icon & Text & Icon &      \\ \midrule
GPT-4o               & -   & 24.9 & 24.0 & 15.5 & 21.4 & 12.2 & 7.3  & 17.8 \\
Qwen2-VL~\cite{qwen2vl}             & 7B  & 78.0 & 62.9 & 64.4 & 50.0 & 72.2 & 61.2 & 66.4 \\
CogAgent~\cite{hong2023cogagent}             & 18B & 68.5 & 21.0 & 73.7 & 17.9 & 70.4 & 33.5 & 49.8 \\
SeeClick~\cite{cheng2024seeclick}             & 10B & 77.3 & 52.4 & 68.6 & 30.7 & 59.1 & 30.6 & 53.4 \\
UGround~\cite{uground}              & 7B  & 82.8 & 61.6 & 84.0 & 65.1 & 81.3 & 71.8 & 74.8 \\
OS-ATLAS-Base~\cite{osatlas}        & 7B  & 93.0 & 72.9 & \textbf{91.8} & 62.9 & \textbf{90.9} & 74.3 & 82.5 \\
UIPro-Qwen2VL (ours) & 7B  & \textbf{93.1} & \textbf{74.7} & 89.2 & \textbf{70.0} & 85.7 & \textbf{74.8} & \textbf{82.7} \\
UIPro-SLiME (ours)   & 3B  & 84.3 & 45.4 & 70.0 & 42.7 & 76.8 & 29.3 & 60.8 \\ \bottomrule
\end{tabular}%
\caption{Grounding accuracy on the subsets of the ScreenSpot benchmark~\cite{cheng2024seeclick}. UIPro leads in the overall performance, and achieves the highest accuracy on icon grounding tasks.}
\label{tab:sspot detail}
\end{table*}

\begin{table*}[]
\centering
\tiny
\begin{tabular}{@{}cccccccccccccccc@{}}
\toprule
\multirow{2}{*}{GUI Source} & \multirow{2}{*}{\#Images} &  \multirow{2}{*}{\#Tasks} & \multicolumn{2}{c}{Text} & \multicolumn{2}{c}{Icon} & \multicolumn{2}{c}{Other Element} & \multicolumn{2}{c}{Functionality} & \multirow{2}{*}{Intent Gnd.} & \multirow{2}{*}{Elem. Class.} & \multirow{2}{*}{Widget Listing} & \multirow{2}{*}{QA} & \multirow{2}{*}{Captioning} \\ \cmidrule(lr){4-11}
                            &                               &                          & Gnd.        & Ref.       & Gnd.        & Ref.       & Gnd.             & Ref.           & Gnd.            & Ref.            &                              &                               &                                 &                     &                             \\ \midrule
Common Crawl\footnote{https://index.commoncrawl.org/}\textsuperscript{\textdagger}                & 35.8k                    & 1.3M                  & 134.2k      & 157.7k     & -           & -          & -                & -              & 314.5k          & 314.5k          & 314.5k                       & -                             & 48.1k                           & -                   & -                           \\
Android Emulator\textsuperscript{\textdagger}            & 29.6k                    & 629.5k                   & 127.5k      & 198.1k     & -           & -          & -                & -              & 38.4k           & 38.4k           & 189.6k                       & -                             & 37.7k                           & -                   & -                           \\
RICO~\cite{deka2017rico}\textsuperscript{*}                        & 34.4k                     & 1.1M                  & -           & -          & -           & -          & 939.5k           & 101.4k         & -               & -               & -                            & -                             & -                               & -                   & 47.2k                       \\
MobileViews~\cite{mobileviews}\textsuperscript{\textdagger}                 & 64.6k                     & 1.6M                  & 497.2k      & 544.3k     & -           & -          & -                & -              & -               & -               & 495.3k                       & -                             & 64.0k                           & -                   & -                           \\
WAE~\cite{chen2020wireframe}\textsuperscript{\textdagger}                         & 372.1k                       & 7.5M                  & 2.1M     & 2.4M    & 241.7k      & 313.5k     & -                & -              & -               & -               & 2.1M                      & -                             & 345.5k                          & -                   & -                           \\
AndroidControl~\cite{androidcontrol}\textsuperscript{\textdagger}              & 23.4k                       & 1.1M                  & 186.3k      & 262.5k     & -           & -          & -                & -              & 23.4k           & 23.4k           & 577.0k                       & -                             & 23.4k                           & -                   & -                           \\
WebUI~\cite{WebUI}\textsuperscript{\textdagger}                        & 162.3k                    & 485.2k                   & -           & -          & 59.6k       & 113.7k    & -                & -              & -               & -               & 110.7k                      & 40.9k                        & 160.3k                        & -                   & -                           \\
MultiUI~\cite{multiUI}\textsuperscript{*}                      & 1.4M                    & 5.2M                 &             & 371.5k    & -           & -          & 684.7k           & -              & -               & -               & 1.2M                     & -                             & -                               & 2.9M            & -                           \\
GUIEnv~\cite{GUICourse}\textsuperscript{*}                       & 70.5k                      & 680.7k                  & 328.3k      & 352.3k   & -           & -          & -                & -              & -               & -               & -                            & -                             & -                               & -                   & -                           \\
SeeClick-Web~\cite{cheng2024seeclick}\textsuperscript{*}                 & 270.8k                     & 1.1M               & 541.5k      & 541.5k    & -           & -          & -                & -              & -               & -               & -                            & -                             & -                               & -                   & -                           \\
OmniAct~\cite{kapoor2024omniact}\textsuperscript{*}                      & 176                                & 19.1k                   & -           & -          & -           & -          & -                & -              & -               & -               & 19.1k                       & -                             & -                               & -                   & -                           \\
MOTIF~\cite{Burns2022ADF}\textsuperscript{*}                        & 55                           & 7.9k                     & -           & -          & -           & -          & -                & -              & -               & -               & 7.9k                        & -                             & -                               & -                   & -                           \\
TOTAL                       & 2.5M                   & 20.6M                & 3.9M    & 4.8M    & 301.3k      & 427.3k    & 1.6M          & 101.4k      & 376.3k          & 376.3k          & 5.0M                      & 40.9k                        & 679.1k                          & 2.9M            & 47.2k                       \\ \bottomrule
\end{tabular}
\caption{\textbf{Data sources and statistics of UI-Pro GUI understanding dataset.} Approximately two thirds of the tasks are newly generate by the authors while one third is cleaned and included from existing dataset. \textdagger means that the authors generate fine-tuning samples from unlabeled raw GUI data. * means that the authors clean the samples provided by the original datasets.}
\label{tab:UI gnd data stats}
\end{table*}

\begin{table}[]
\centering
\begin{tabular}{@{}cc@{}}
\toprule
GUI Source       & \%Invalid Elem \\ \midrule
Common Crawl     & 2.5            \\
Android Emulator & 0.4            \\
MobileViews~\cite{mobileviews}      & 24.5           \\
WAE~\cite{chen2020wireframe}              & 29.0           \\
AndroidControl~\cite{androidcontrol}   & 11.5           \\
WebUI~\cite{WebUI}            & 14.4           \\
MultiUI~\cite{multiUI}          & 3.0            \\
SeeClick-Web~\cite{cheng2024seeclick}     & 0.2            \\ \bottomrule
\end{tabular}
\caption{The percentage of invalid elements detected for the used data source by the proposed denoising procedure.}
\label{tab:noise ratio}
\end{table}

\begin{table}[]
\scriptsize
\centering
\begin{tabular}{@{}ccccc@{}}
\toprule
Scenario                & Data Source   & \#Used Steps & Total                   \\ \midrule
\multirow{6}{*}{Mobile} & AITW~\cite{rawles2023android}               & 37.6k     & \multirow{6}{*}{380.0k} \\
                        & AITZ~\cite{AITZ}       & 25.9k     &                         \\
                        & AMEX~\cite{AMEX}          & 38.7k     &                         \\
                        & AndroidControl~\cite{androidcontrol}   & 124.0k    &                         \\
                        & GUIAct-smartphone~\cite{GUICourse}  & 64.3k     &                         \\
                        & GUIOdyssey~\cite{GUIOdyssey}      & 107.7k    &                         \\ \midrule
\multirow{3}{*}{Web}    & Mind2Web~\cite{deng2024mind2web}        & 7.7k      & \multirow{3}{*}{144.9k} \\
                        & GUIAct-web~\cite{GUICourse}        & 109.3k                         \\
                        & WebLINX~\cite{weblinx}           & 22.0k                         \\ \bottomrule
\end{tabular}
\caption{The number of samples included in the merged GUI agen task fine-tuning data from each data source.}
\label{tab:agent tasks}
\end{table}

\begin{figure*}[ht]
    \centering
    \includegraphics[width=1\linewidth]{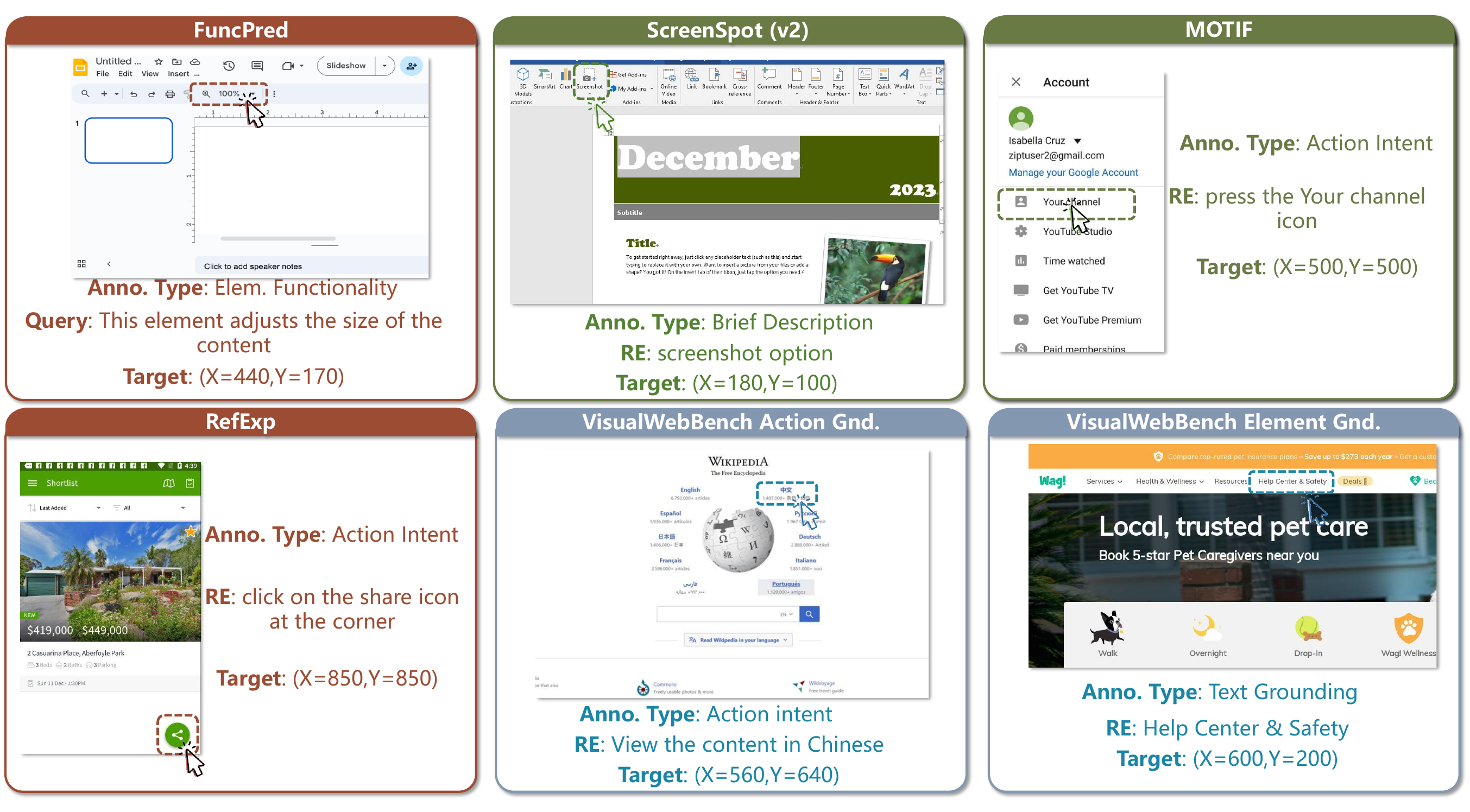}
    \caption{Examples of the grounding tasks provided by the used GUI element grounding benchmarks in Sec.\ref{sec:gnd evaluation}.}
    \label{fig:gnd example}
\end{figure*}

\end{document}